%% file: neurips_2026.tex
\documentclass{article}

\usepackage[preprint]{neurips_2026}

\usepackage[utf8]{inputenc}
\usepackage[T1]{fontenc}
\usepackage{float}
\usepackage{hyperref}
\usepackage{url}
\usepackage{booktabs}
\usepackage{amsfonts}
\usepackage{amsmath}
\usepackage{amssymb}
\usepackage{nicefrac}
\usepackage{microtype}
\usepackage{xcolor}
\usepackage{graphicx}
\usepackage{multirow}
\hypersetup{hidelinks}

\title{MRFlow: Mechanism-Structured Residual Flow Matching for Virtual Cell Perturbation Prediction}

\author{Haojie Yang, Ran Su$^*$\\
College of Intelligence and Computing, Tianjin University, Tianjin, China\\
\texttt{haojae@tju.edu.cn}, \texttt{ran.su@tju.edu.cn}\\
$^*$Corresponding author}

\begin{document}
\raggedbottom
\setlength{\parskip}{2.2pt}
\setlength{\intextsep}{6pt plus 1pt minus 1pt}
\setlength{\textfloatsep}{8pt plus 1pt minus 2pt}
\setlength{\floatsep}{6pt plus 1pt minus 1pt}
\setlength{\abovedisplayskip}{3pt plus 1pt minus 1pt}
\setlength{\belowdisplayskip}{3pt plus 1pt minus 1pt}
\setlength{\abovedisplayshortskip}{2pt plus 1pt minus 1pt}
\setlength{\belowdisplayshortskip}{2pt plus 1pt minus 1pt}

\maketitle

\begin{figure}[H]
  \centering
  \includegraphics[width=\linewidth]{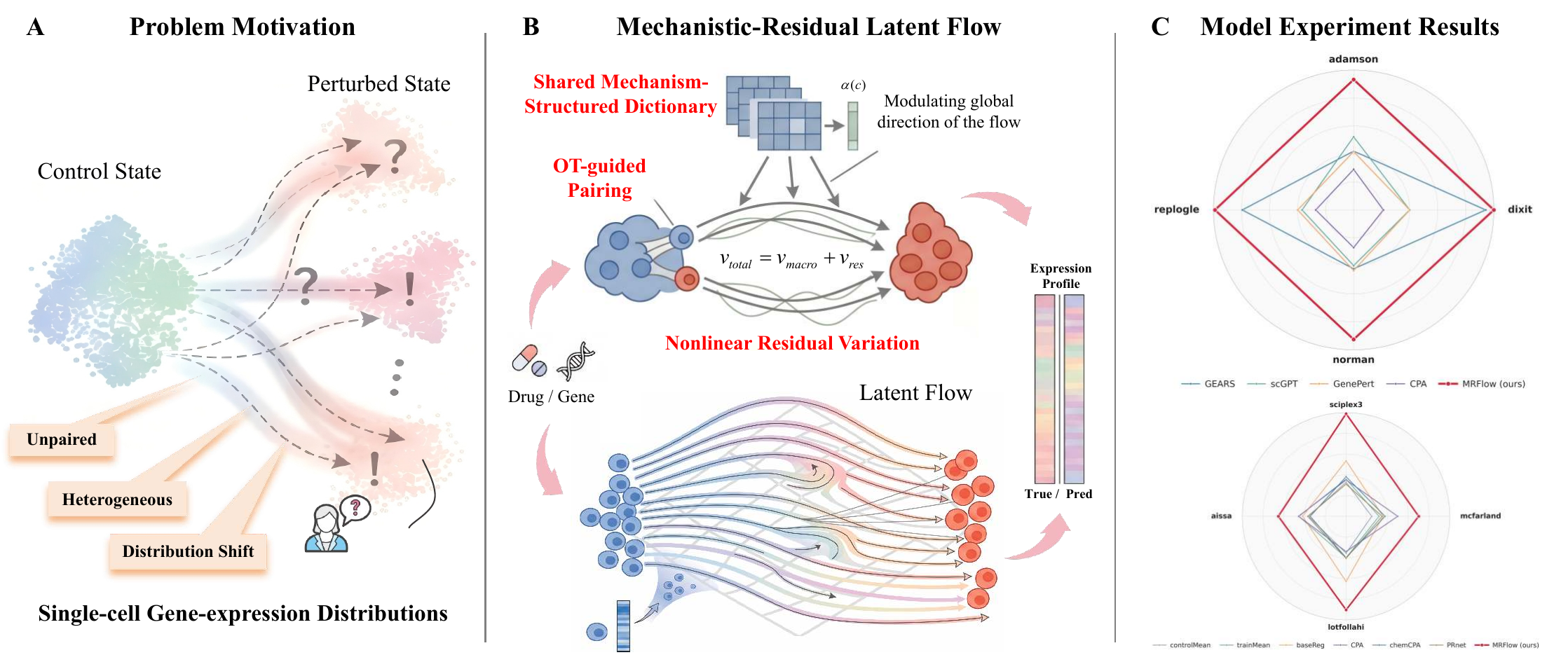}
  \caption{MRFlow overview. Unpaired control and perturbed cells are linked through condition-specific latent transport, with a mechanism-structured affine field, a controlled residual field, and evaluation across gene and drug perturbation benchmarks.}
  \label{fig:overview}
\end{figure}

\begin{abstract}
Predicting cellular responses to chemical and genetic perturbations is a central test for AI virtual cells, yet many single-cell models reduce each condition to an average expression shift. This target is misaligned with perturbation biology: rare states, resistant subpopulations, and context-dependent effect sizes may coexist under the same intervention. MRFlow treats virtual-cell prediction as conditional transport between empirical single-cell measures. Control and perturbed cells are embedded in a variational latent state space, greedy low-cost supervision is constructed over detached minibatch states, and a conditional flow-matching vector field transports control populations toward perturbed populations. The field decomposes into a mechanism-structured affine dictionary, which recombines shared latent response operators with a low-rank condition-specific correction, and a spectrally controlled residual field for nonlinear cell-level variation. Genetic perturbations are encoded through a regulatory-graph GCN, chemical perturbations through Morgan fingerprints, and both enter the same conditional dynamics interface. Across eight gene and drug perturbation benchmarks, MRFlow attains the highest composite score while preserving competitive raw performance on differential-expression and distributional metrics. The results support latent distribution transport as a principled and empirically effective formulation for perturbation-aware virtual cell modeling.
\end{abstract}

\newpage
\section{Introduction}
Accurate prediction of single-cell responses to interventions is a concrete test of whether AI virtual cells can move beyond transcriptomic annotation toward predictive perturbation modeling. Large Perturb-seq and chemical transcriptomics screens now measure the transcriptional effects of genetic and pharmacological perturbations at single-cell resolution \citep{dixit2016perturbseq,adamson2016multiplexed,norman2019exploring,replogle2022mapping,srivatsan2020massively,mcfarland2020multiplexed}, and recent resources have harmonized these measurements for broader comparison \citep{peidli2024scperturb,wei2025perturbase}. These atlases provide the empirical basis for virtual cell models that anticipate held-out perturbation conditions, guide functional genomics, and narrow the experimental search space in drug discovery \citep{ji2021machine,bunne2024virtual,rood2024toward}.

The dominant computational abstraction remains conditional response regression. scGen introduced latent arithmetic for out-of-sample perturbation prediction \citep{lotfollahi2019scgen}; CPA and chemCPA extended conditional autoencoding to compositional treatments, dose response, and unseen drugs \citep{lotfollahi2023predicting,hetzel2022predicting}; GEARS and TxPert used graph-structured biological knowledge to predict transcriptional effects of genetic perturbations \citep{roohani2024predicting,wenkel2026txpert}; and representation models such as biolord, scGPT, and scFoundation have been used for perturbation prediction or disentangled response analysis \citep{piran2024disentanglement,cui2024scgpt,hao2024large}. These systems have expanded the modeling toolkit, but the object they are often asked to predict is still a condition-level mean transcriptional effect.

The mean-shift view misses a defining feature of single-cell perturbation biology. A drug or gene perturbation induces a response distribution, in which resistant subpopulations, rare transcriptional states, and cell-state-dependent effect sizes carry biologically meaningful signal \citep{replogle2022mapping,srivatsan2020massively,aissa2021single}. A second limitation is exposed by recent benchmarks: simple baselines, systematic variation, and drug-response evaluation context can all change the apparent ranking of perturbation predictors \citep{ahlmann2025deep,wong2025simple,vinas2025systema,wei2026benchmarking,wang2026scdrugmap}. These results suggest that additional capacity alone is an insufficient answer. A virtual cell model needs a hypothesis class that can generate response distributions while controlling how perturbation dynamics extrapolate to new conditions.

Flow matching and optimal transport provide a mathematical framework for this distributional problem. Flow matching trains continuous normalizing flows by regressing vector fields along probability paths, and OT-guided variants reduce transport cost and path crossing through minibatch couplings \citep{lipman2023flow,liu2023rectified,tong2024improving}. In single-cell biology, optimal transport has been used for developmental alignment, data integration, cellular dynamics, and perturbation response modeling \citep{schiebinger2019optimal,cao2022unified,tong2022trajectorynet,bunne2023learning,bunne2024optimal}. Standard flow models, however, do not by themselves address the unpaired nature of perturbation data or the need for structured extrapolation across conditions.

MRFlow instantiates this view by learning perturbation-conditioned transport dynamics in a variational latent cell-state space, as outlined in Figure~\ref{fig:overview}. The model maps a control-cell distribution to a perturbed-cell distribution through OT-guided flow matching. Its vector field is not an unconstrained neural network: the dominant component is a mechanism-structured affine dictionary, in which a condition-specific matrix is assembled from shared response operators plus a low-rank correction. A spectrally regularized residual field captures nonlinear cell-level variation that remains after the affine component. Genetic and chemical interventions enter through modality-specific encoders, but both parameterize the same latent dynamics. A negative binomial decoder maps transported latent states back to count-aware expression distributions.

The empirical study spans eight human single-cell perturbation datasets covering CRISPR activation, CRISPR interference, CRISPR knockout, multiplexed drug screens, multi-dose drug response, and combination treatments. Across these datasets, MRFlow obtains the highest composite score against statistical baselines, conditional autoencoders, graph models, chemical-response models, gene-embedding linear models, and a single-cell foundation model. Ablations attribute performance to the mechanism-structured affine field, residual field, low-rank correction, OT-guided pairing, and count-aware training. Operator-level analyses further show response patterns consistent with pathway-level perturbation groups.

\section{Related Work}

\subsection{Single-cell perturbation prediction}

Single-cell perturbation prediction has developed alongside experimental platforms that pair pooled interventions with scRNA-seq readouts. Perturb-seq enabled scalable CRISPR-linked molecular profiling \citep{dixit2016perturbseq,adamson2016multiplexed,norman2019exploring,replogle2022mapping}, while sci-Plex, MIX-Seq, scPerturb, and PerturBase extended chemical screens and reusable perturbation resources \citep{srivatsan2020massively,mcfarland2020multiplexed,peidli2024scperturb,wei2025perturbase}. Computational models built on these data include scGen \citep{lotfollahi2019scgen}, CPA \citep{lotfollahi2023predicting}, chemCPA \citep{hetzel2022predicting}, PRnet \citep{qi2024predicting}, GEARS \citep{roohani2024predicting}, TxPert \citep{wenkel2026txpert}, GenePert \citep{chen2024genepert}, biolord \citep{piran2024disentanglement}, and PerturbNet \citep{yu2025perturbnet}. Recent benchmark studies stress that split design and systematic variation materially affect conclusions about generalization \citep{vinas2025systema,wei2026benchmarking}. MRFlow takes a different route: it models the target as a conditional response distribution and constrains the latent vector field through a shared affine dictionary.

\subsection{Single-cell foundation models}

Foundation models trained on large transcriptomic corpora are a natural reference point for virtual cell modeling. Geneformer demonstrated transfer learning for gene network biology \citep{theodoris2023transfer}, scGPT introduced generative pretraining across tens of millions of cells \citep{cui2024scgpt}, and scFoundation scaled masked expression modeling with an asymmetric transformer architecture \citep{hao2024large}. These models provide broad representations, while task-specific perturbation benchmarks show that representation scale does not automatically translate into stronger held-out perturbation prediction \citep{ahlmann2025deep,wong2025simple,wei2026benchmarking,wang2026scdrugmap}. MRFlow is complementary to this line of work: it models the perturbation response operator directly and constrains the conditional dynamics instead of relying on pretraining scale.

\subsection{Flow matching and optimal transport}

Continuous normalizing flows define generative models through neural ODEs \citep{chen2018neural}. Flow matching, rectified flow, and stochastic interpolants train these dynamics without repeatedly solving likelihood-based ODE objectives during training \citep{lipman2023flow,liu2023rectified,albergo2023stochastic}. Minibatch OT improves flow paths by coupling source and target samples through low-cost assignments \citep{tong2024improving}. Optimal transport has also become a core tool in single-cell analysis, including developmental trajectory reconstruction, integration, dynamics, and perturbation response \citep{schiebinger2019optimal,cao2022unified,tong2022trajectorynet,bunne2023learning,bunne2024optimal,klein2023moscot}. MRFlow adapts these ideas to latent virtual cell prediction, where the main obstacles are unpaired perturbation measurements and generalization across held-out conditions.

\subsection{Generative count models and perturbation encoders}

Deep generative models for scRNA-seq commonly use negative binomial likelihoods to account for library size and overdispersion \citep{lopez2018deep,xu2021probabilistic,gayoso2022totalvi}. MRFlow adopts this count-aware view but inserts a conditional transport model between encoding and decoding. For perturbation conditioning, prior work has used gene regulatory or ontology graphs for genetic interventions \citep{roohani2024predicting}, curated transcription-factor networks such as CollecTRI \citep{muller2024collectri}, and molecular fingerprints for chemical structure \citep{rogers2010extended,hetzel2022predicting}. MRFlow uses separate gene and drug encoders only up to the condition embedding; after that point, both modalities control the same mechanism-structured residual vector field.

\section{Method}

\begin{figure}[t]
  \centering
  \includegraphics[width=0.96\linewidth]{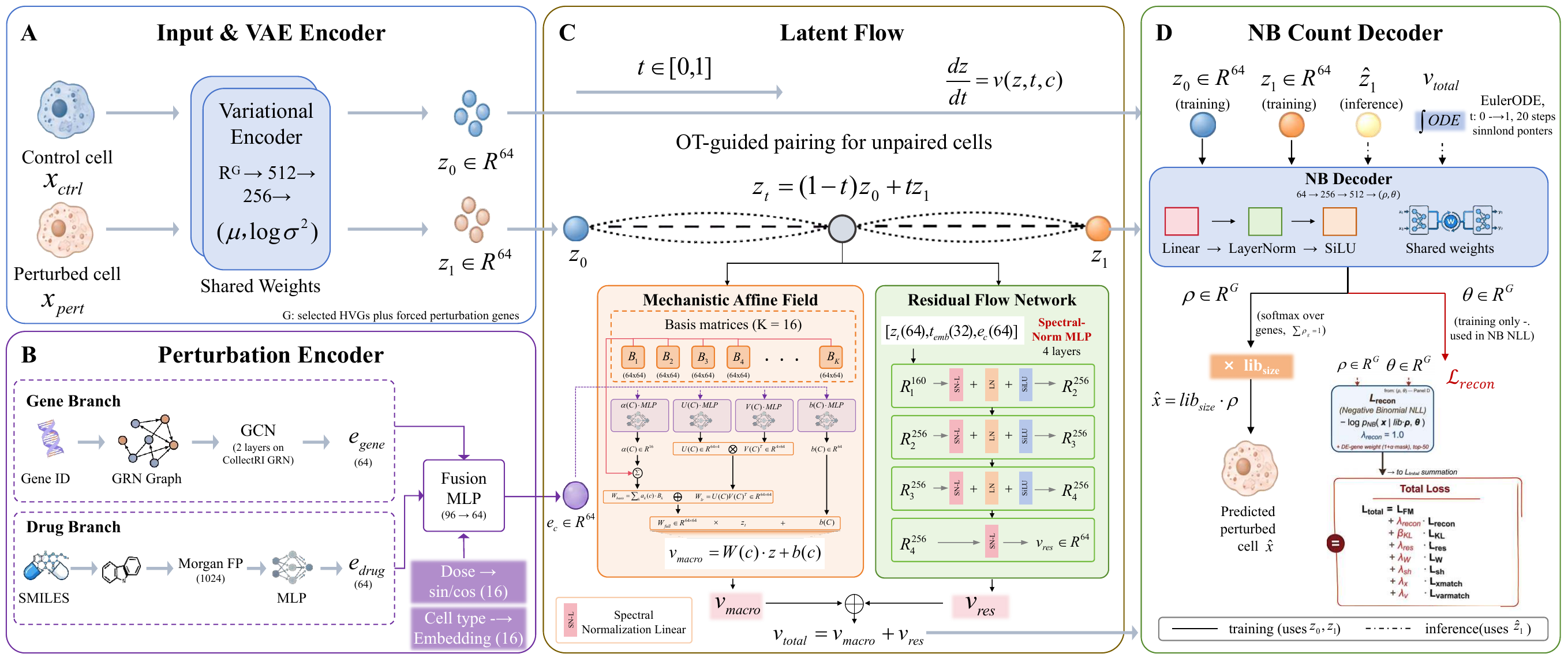}
  \caption{MRFlow architecture. Variational cell-state encoding, modality-specific perturbation encoding, low-cost latent pairing, mechanism-structured residual flow matching, and negative binomial decoding form a count-aware virtual cell predictor.}
  \label{fig:architecture}
\end{figure}

\subsection{Virtual cell prediction as latent distribution transport}

Single-cell perturbation data provide empirical populations rather than paired before-and-after observations. For a perturbation condition \(c\), let \(\mathcal{X}_0=\{x_i^0\}_{i=1}^{n_0}\) denote control cells and \(\mathcal{X}_c=\{x_j^c\}_{j=1}^{n_c}\) denote perturbed cells. Each \(x\in\mathbb{R}^{G}\) is a log-transformed expression vector over \(G\) genes. The count-scale target is \(y\in\mathbb{N}^{G}\), and the scalar \(\ell\) denotes the cell-specific library size. The condition \(c=(a,s,r)\) contains perturbation identity \(a\), dose \(s\), and cell context \(r\). Genetic perturbations use \(s=1\) when dose is not defined.

MRFlow operates in a continuous latent cell-state space of dimension \(d\). A variational encoder \(E_\phi\) maps expression to a Gaussian posterior,
\begin{equation}
q_\phi(z\mid x)=
\mathcal{N}\left(\mu_\phi(x),\operatorname{diag}\left(\sigma_\phi^2(x)\right)\right),
\label{eq:encoder}
\end{equation}
where \(z\in\mathbb{R}^{d}\) is the latent cell state, and \(\mu_\phi(x)\) and \(\sigma_\phi(x)\) are encoder outputs. During training, \(z=\mu_\phi(x)+\sigma_\phi(x)\odot\epsilon\) with \(\epsilon\sim\mathcal{N}(0,I_d)\). During inference, \(z=\mu_\phi(x)\).

The encoder induces latent control and perturbed distributions,
\begin{equation}
\mathbb{Q}_0^\phi=E_\phi\#\mathbb{P}_0,
\qquad
\mathbb{Q}_c^\phi=E_\phi\#\mathbb{P}_c,
\label{eq:pushforward}
\end{equation}
where \(\mathbb{P}_0\) and \(\mathbb{P}_c\) are empirical expression-space distributions, and \(\#\) denotes distribution pushforward. The virtual cell objective is to learn a condition-specific flow map \(\Phi_c:\mathbb{R}^{d}\rightarrow\mathbb{R}^{d}\) such that
\begin{equation}
\Phi_c\#\mathbb{Q}_0^\phi \approx \mathbb{Q}_c^\phi .
\label{eq:transport}
\end{equation}
This formulation makes the prediction target a population distribution rather than a single response vector.

\subsection{OT-guided latent flow matching for unpaired data}

Perturbation experiments do not identify which control cell should correspond to which perturbed cell. Given a minibatch of encoded control states \(Z_0=\{z_i^0\}_{i=1}^{N}\) and encoded perturbed endpoint states \(Z_1=\{z_j^1\}_{j=1}^{N}\), MRFlow constructs a low-cost latent pairing. The squared Euclidean cost between a control state and a perturbed endpoint is computed over detached latent states,
\begin{equation}
C_{ij}=\left\|\operatorname{stopgrad}(z_i^0)-\operatorname{stopgrad}(z_j^1)\right\|_2^2 .
\label{eq:cost}
\end{equation}
The default implementation uses greedy nearest-neighbor pairing in this latent geometry and denotes the paired endpoint for \(z_i^0\) by \(z_{1,i}^{\pi}\). Pairing is recomputed for each minibatch and is not differentiated; implementation details are given in Appendix~\ref{app:pairing}. This construction supplies transport-guided supervision without asserting that each minibatch solves exact global OT.

For each paired latent endpoint pair \((z_0,z_1^\pi)\), MRFlow samples \(t\sim\mathcal{U}(0,1)\) and constructs the linear interpolation
\begin{equation}
z_t=(1-t)z_0+t z_1^\pi .
\label{eq:path}
\end{equation}
The target velocity along this path is the displacement \(u_t=z_1^\pi-z_0\). The model learns a conditional vector field \(v_\Theta(z,t,c)\) through flow matching,
\begin{equation}
\mathcal{L}_{FM}=
\mathbb{E}_{c,z_0,z_1^\pi,t}
\left[
\left\|v_\Theta(z_t,t,c)-(z_1^\pi-z_0)\right\|_2^2
\right].
\label{eq:fm}
\end{equation}
Low-cost pairing reduces the noisy endpoint assignments that arise when unpaired cell populations are matched at random.

\subsection{Mechanism-structured residual vector field}

The central architectural choice in MRFlow is the decomposition of the total velocity field into a structured affine component and a neural residual component, as shown in Figure~\ref{fig:architecture}:
\begin{equation}
v_\Theta(z,t,c)=v_{\mathrm{macro}}(z,c)+v_{\mathrm{res}}(z,t,c).
\label{eq:decomp}
\end{equation}
The term \(v_{\mathrm{macro}}\) parameterizes the dominant condition-specific transport trend, while \(v_{\mathrm{res}}\) models nonlinear deviations and cell-level heterogeneity.

The macro field is affine in the latent state,
\begin{equation}
v_{\mathrm{macro}}(z,c)=W(c)z+b(c),
\label{eq:macro}
\end{equation}
where \(W(c)\in\mathbb{R}^{d\times d}\) is a condition-specific affine matrix and \(b(c)\in\mathbb{R}^{d}\) is a condition-specific bias. Rather than generating \(W(c)\) as an unconstrained dense matrix, MRFlow constrains it to a shared affine dictionary plus a low-rank correction:
\begin{equation}
W(c)=\sum_{k=1}^{K}\alpha_k(e_c)B_k+U(e_c)V(e_c)^\top .
\label{eq:dictionary}
\end{equation}
Here \(e_c\in\mathbb{R}^{d_c}\) is the condition embedding, \(K\) is the number of shared bases, \(B_k\in\mathbb{R}^{d\times d}\) is the \(k\)-th latent response operator, \(\alpha_k(e_c)\) is its condition-specific mixing coefficient, and \(U(e_c),V(e_c)\in\mathbb{R}^{d\times r}\) define a rank-at-most-\(r\) correction. This parameterization replaces arbitrary condition-specific Jacobians with recombinations of shared response modes plus restricted adaptation, giving the vector field a low-dimensional operator prior across conditions.

The residual field is a spectral-normalized neural network,
\begin{equation}
v_{\mathrm{res}}(z,t,c)=g_\omega\left([z,\gamma(t),e_c]\right),
\label{eq:residual}
\end{equation}
where \(g_\omega\) is an MLP, \([\cdot]\) denotes concatenation, and \(\gamma(t)\) is a sinusoidal embedding of flow time. Spectral normalization and an explicit residual norm penalty bound the role of nonlinear corrections. This design encourages the affine dictionary to explain low-order response structure, while the residual field absorbs nonlinear population variation that a single affine operator cannot express.

\subsection{Perturbation-conditioned encoding}

The condition embedding \(e_c\) represents both genetic and chemical interventions. MRFlow uses modality-specific encoders up to a shared condition interface,
\begin{equation}
e_c=f_\psi(a,s,r),
\label{eq:condition}
\end{equation}
where \(f_\psi\) fuses perturbation identity, dose, and cell context. For gene perturbations, a gene regulatory graph \(\mathcal{G}=(\mathcal{V},\mathcal{E})\) with adjacency matrix \(A\) provides the message-passing structure. If \(H^{(0)}\in\mathbb{R}^{G\times d_c}\) denotes initial gene embeddings, a graph convolutional layer is
\begin{equation}
H^{(m+1)}=
\sigma\left(\tilde{D}^{-1/2}\tilde{A}\tilde{D}^{-1/2}H^{(m)}R^{(m)}\right),
\label{eq:gcn}
\end{equation}
where \(\tilde{A}=A+I\), \(\tilde{D}_{ii}=\sum_j\tilde{A}_{ij}\), \(R^{(m)}\) is a trainable weight matrix, and \(\sigma\) is a nonlinear activation. A single-gene perturbation \(a=g\) uses \(e_a=H_g^{(L)}\). For a two-gene perturbation \(a=(g_1,g_2)\), MRFlow combines additive and interaction terms,
\begin{equation}
e_a=H_{g_1}^{(L)}+H_{g_2}^{(L)}
+\psi_{\mathrm{int}}\left(H_{g_1}^{(L)}\odot H_{g_2}^{(L)}\right),
\label{eq:combo}
\end{equation}
where \(\odot\) denotes elementwise product and \(\psi_{\mathrm{int}}\) is a learned interaction network.

For drug perturbations, the chemical identity \(a\) is represented by a Morgan fingerprint \(m(a)\in\{0,1\}^{F}\), which an MLP maps to \(e_a=\psi_{\mathrm{drug}}(m(a))\). Dose is encoded on a log scale through sinusoidal features, and cell context is represented by a learned embedding. The fused vector \(e_c\) controls \(\alpha(e_c)\), \(U(e_c)\), \(V(e_c)\), \(b(c)\), and the residual field in Equation~\ref{eq:residual}.

\subsection{Count-aware decoding and full objective}

MRFlow decodes latent states through a negative binomial count model. The decoder outputs gene proportions and dispersion parameters,
\begin{equation}
\rho_\theta(z)=\operatorname{softmax}(h_\theta(z)),
\qquad
\theta_\theta(z)=\exp(r_\theta(z)),
\label{eq:decoder}
\end{equation}
where \(\rho_{\theta,g}(z)\) is the predicted proportion of gene \(g\), \(\sum_{g=1}^{G}\rho_{\theta,g}(z)=1\), and \(\theta_{\theta,g}(z)>0\) is the dispersion for gene \(g\). For cell \(i\) with library size \(\ell_i\), the count mean for gene \(g\) is
\begin{equation}
\mu_{ig}=\ell_i\rho_{\theta,g}(z_i),
\qquad
y_{ig}\sim \operatorname{NB}(\mu_{ig},\theta_{\theta,g}(z_i)).
\label{eq:nb}
\end{equation}
The reconstruction loss \(\mathcal{L}_{recon}\) is the negative log likelihood under this count model for control and perturbed cells.

The training objective combines flow matching, count reconstruction, variational regularization, structured-field regularization, and auxiliary entropic OT alignment:
\begin{equation}
\mathcal{L}=
\mathcal{L}_{FM}
+\lambda_{recon}\mathcal{L}_{recon}
+\beta_{KL}\mathcal{L}_{KL}
+\lambda_{res}\mathcal{L}_{res}
+\lambda_W\mathcal{L}_{W}
+\lambda_{OT}\mathcal{L}_{OT}^{aux}.
\label{eq:objective}
\end{equation}
The KL term regularizes \(q_\phi(z\mid x)\) toward \(\mathcal{N}(0,I_d)\). The residual penalty is \(\mathcal{L}_{res}=\mathbb{E}\|v_{\mathrm{res}}(z_t,t,c)\|_2^2\), and the affine penalty is \(\mathcal{L}_{W}=\mathbb{E}_{c}\|W(c)\|_F^2\). The auxiliary OT term compares predicted endpoints with paired perturbed endpoints through an entropic transport cost. Appendix~\ref{app:objective} gives stabilization details and the precise interpretation of each term.

\subsection{Inference}

At inference time, MRFlow starts from control cells and integrates the learned conditional ODE. For a control cell \(x_0\), the initial latent state is \(z_0=\mu_\phi(x_0)\). The predicted latent endpoint is
\begin{equation}
\hat{z}_1=\Phi_c(z_0)=z_0+\int_0^1 v_\Theta(z_t,t,c)\,dt,
\qquad
\frac{dz_t}{dt}=v_\Theta(z_t,t,c).
\label{eq:inference}
\end{equation}
Applying Equation~\ref{eq:inference} to a population of control cells yields a predicted perturbed-cell distribution. The NB decoder then produces count means, which are transformed into the expression space used for evaluation.

\section{Experiments}

\subsection{Datasets, baselines, and metrics}

The evaluation covers four genetic perturbation datasets and four drug perturbation datasets. The genetic benchmarks are Norman CRISPRa in K562 cells, Replogle CRISPRi in RPE1 cells, Adamson CRISPRi in K562 cells, and Dixit CRISPR knockout in K562 cells \citep{norman2019exploring,replogle2022mapping,adamson2016multiplexed,dixit2016perturbseq}. The drug benchmarks are SciPlex3 across MCF7, A549, and K562 cells, McFarland MIX-Seq across 169 cancer cell lines, Aissa in PC9 cells, and the Lotfollahi A549 drug-combination dataset \citep{srivatsan2020massively,mcfarland2020multiplexed,aissa2021single,lotfollahi2023predicting}. Together, these datasets contain 1.50 million human single-cell profiles and 2,121 perturbation conditions.

MRFlow is compared with deterministic baselines controlMean, trainMean, and baseReg \citep{ahlmann2025deep,wong2025simple}; conditional autoencoders CPA \citep{lotfollahi2023predicting} and chemCPA \citep{hetzel2022predicting}; PRnet \citep{qi2024predicting}; graph-based GEARS \citep{roohani2024predicting}; GenePert \citep{chen2024genepert}; and scGPT \citep{cui2024scgpt}. Learned baselines use official open-source implementations, while controlMean, trainMean, and baseReg are deterministic benchmark rules. Within each dataset, every method uses the same condition-level train and test split, preprocessed expression matrix, benchmark vocabulary, and evaluation script. Metrics include DE20 Pearson, DE20 MSE, median Wasserstein distance, and E-distance. The composite score applies per-dataset min-max normalization followed by an equal-weight average over the four normalized metrics. Appendix~\ref{app:datasets} and Appendix~\ref{app:baseline} give dataset, split, preprocessing, and baseline-alignment details.

MRFlow provides a unified architecture and condition interface for genetic and chemical perturbations. In these benchmarks, models are trained and evaluated per dataset or task using the same MRFlow design, rather than as one jointly trained cross-dataset model.

\subsection{Main results}

MRFlow ranks first by composite score on all eight benchmarks, and the ranking is supported by raw metrics rather than by normalization alone. On SciPlex3, MRFlow improves Pearson and MSE over baseReg, CPA, chemCPA, PRnet, trainMean, and controlMean, while also achieving the lowest E-distance. On Replogle, GenePert has slightly higher Pearson than MRFlow, but MRFlow has much lower Wasserstein distance and E-distance, indicating better population-level agreement. Tables~\ref{tab:drug_raw} and~\ref{tab:gene_raw} report the full raw metrics; Appendices~\ref{app:rawmetrics}, \ref{app:composite}, and~\ref{app:add_quant} provide modality-level visualizations, the composite-score definition, and rank-based diagnostics over raw metric cells.

\begin{table}[H]
  \caption{Drug perturbation raw metrics on held-out conditions. Pearson and MSE use DE20 genes.}
  \label{tab:drug_raw}
  \centering
  \setlength{\abovecaptionskip}{0pt}
  \setlength{\belowcaptionskip}{0pt}
  \scriptsize
  \renewcommand{\arraystretch}{0.58}
  \setlength{\tabcolsep}{0.8pt}
  \begin{tabular*}{\linewidth}{@{\extracolsep{\fill}}llrrrrrr@{}}
    \toprule
    Dataset & Method & Pearson & MSE & Wass. & E-dist. & Comp. & Rank \\
    \midrule
    \multirow{7}{*}{SciPlex3} & MRFlow & \textbf{0.9608} & \textbf{0.6957} & 0.0002 & \textbf{9.9167} & \textbf{0.9919} & \textbf{1} \\
     & baseReg & 0.8774 & 0.7751 & 0.0004 & 10.9499 & 0.7370 & 2 \\
     & CPA & 0.8712 & 1.4577 & \textbf{0.0001} & 12.5426 & 0.3774 & 5 \\
     & chemCPA & 0.7789 & 1.4426 & 0.0002 & 10.3624 & 0.4545 & 3 \\
     & PRnet & 0.8976 & 1.4240 & 0.0032 & 11.4727 & 0.2764 & 7 \\
     & trainMean & 0.8991 & 1.4195 & 0.0005 & 12.5336 & 0.3969 & 4 \\
     & controlMean & 0.8902 & 1.4110 & 0.0013 & 12.5482 & 0.3215 & 6 \\
    \cmidrule(lr){1-8}
    \multirow{7}{*}{McFarland} & MRFlow & \textbf{0.8909} & \textbf{0.3647} & \textbf{0.0079} & 17.5791 & \textbf{0.7500} & \textbf{1} \\
     & baseReg & 0.8429 & 2.1609 & 0.0145 & 13.6811 & 0.3927 & 4 \\
     & CPA & 0.8551 & 2.1910 & 0.0086 & 9.7808 & 0.7021 & 2 \\
     & chemCPA & 0.8224 & 2.2569 & 0.0161 & \textbf{9.1354} & 0.4383 & 3 \\
     & PRnet & 0.7740 & 2.2813 & 0.0126 & 12.6646 & 0.3683 & 5 \\
     & trainMean & 0.8295 & 2.4655 & 0.0154 & 14.2741 & 0.2941 & 6 \\
     & controlMean & 0.7166 & 2.5805 & 0.0152 & 14.4684 & 0.1195 & 7 \\
    \cmidrule(lr){1-8}
    \multirow{7}{*}{Lotfollahi} & MRFlow & \textbf{0.8866} & 0.5534 & 0.0003 & \textbf{8.1645} & \textbf{0.9193} & \textbf{1} \\
     & baseReg & 0.8817 & \textbf{0.1509} & 0.0003 & 12.1572 & 0.8090 & 2 \\
     & CPA & 0.7946 & 1.0794 & 0.0047 & 13.6452 & 0.3505 & 7 \\
     & chemCPA & 0.8244 & 1.2844 & 0.0021 & 11.4361 & 0.5037 & 5 \\
     & PRnet & 0.8578 & 0.3058 & 0.0132 & 11.0821 & 0.5455 & 3 \\
     & trainMean & 0.7112 & 0.6246 & \textbf{0.0002} & 11.2938 & 0.5145 & 4 \\
     & controlMean & 0.7957 & 1.4278 & 0.0013 & 12.7328 & 0.3909 & 6 \\
    \cmidrule(lr){1-8}
    \multirow{7}{*}{Aissa} & MRFlow & 0.9814 & 0.6060 & \textbf{0.0095} & \textbf{6.2952} & \textbf{0.7109} & \textbf{1} \\
     & baseReg & 0.9789 & 0.2149 & 0.0123 & 8.7354 & 0.5909 & 4 \\
     & CPA & 0.9735 & 0.0511 & 0.0194 & 7.0187 & 0.6563 & 2 \\
     & chemCPA & 0.9652 & \textbf{0.0317} & 0.0279 & 7.9376 & 0.3637 & 7 \\
     & PRnet & 0.9696 & 0.2759 & 0.0289 & 6.5901 & 0.4242 & 5 \\
     & trainMean & \textbf{0.9844} & 0.1889 & 0.0145 & 8.6636 & 0.6520 & 3 \\
     & controlMean & 0.9707 & 0.2007 & 0.0167 & 9.0472 & 0.4053 & 6 \\
    \bottomrule
  \end{tabular*}
\end{table}

\begin{table}[H]
  \caption{Genetic perturbation raw metrics on held-out conditions. Pearson and MSE use DE20 genes.}
  \label{tab:gene_raw}
  \centering
  \setlength{\abovecaptionskip}{0pt}
  \setlength{\belowcaptionskip}{0pt}
  \scriptsize
  \renewcommand{\arraystretch}{0.58}
  \setlength{\tabcolsep}{0.8pt}
  \begin{tabular*}{\linewidth}{@{\extracolsep{\fill}}llrrrrrr@{}}
    \toprule
    Dataset & Method & Pearson & MSE & Wass. & E-dist. & Comp. & Rank \\
    \midrule
    \multirow{5}{*}{Adamson} & MRFlow & \textbf{0.9210} & \textbf{0.1420} & 0.0418 & \textbf{2.2312} & \textbf{0.9434} & \textbf{1} \\
     & GEARS & 0.8986 & 0.1750 & 0.0715 & 2.5053 & 0.6624 & 3 \\
     & GenePert & 0.9139 & 0.1552 & 0.0485 & 10.1359 & 0.6523 & 4 \\
     & scGPT & 0.9157 & 0.2057 & \textbf{0.0331} & 7.3257 & 0.7944 & 2 \\
     & CPA & 0.8162 & 0.4529 & 0.0370 & 11.2168 & 0.2246 & 5 \\
    \cmidrule(lr){1-8}
    \multirow{5}{*}{Dixit} & MRFlow & \textbf{0.9630} & \textbf{0.0158} & \textbf{0.0449} & \textbf{0.5232} & \textbf{1.0000} & \textbf{1} \\
     & GEARS & 0.9591 & 0.0165 & 0.0463 & 0.5669 & 0.9833 & 2 \\
     & GenePert & 0.9335 & 0.0314 & 0.0666 & 5.7759 & 0.5314 & 3 \\
     & scGPT & 0.8449 & 0.0593 & 0.0587 & 4.3082 & 0.5272 & 4 \\
     & CPA & 0.5837 & 0.1110 & 0.0792 & 5.1365 & 0.0304 & 5 \\
    \cmidrule(lr){1-8}
    \multirow{5}{*}{Norman} & MRFlow & \textbf{0.9715} & 0.0940 & 0.0048 & \textbf{0.8123} & \textbf{0.9381} & \textbf{1} \\
     & GEARS & 0.9542 & 0.1128 & 0.0198 & 0.8651 & 0.6634 & 3 \\
     & GenePert & 0.9685 & \textbf{0.0869} & 0.0056 & 6.1470 & 0.6904 & 2 \\
     & scGPT & 0.9211 & 0.1885 & \textbf{0.0006} & 6.1124 & 0.4950 & 4 \\
     & CPA & 0.8971 & 0.3341 & 0.0066 & 6.5010 & 0.1719 & 5 \\
    \cmidrule(lr){1-8}
    \multirow{5}{*}{Replogle} & MRFlow & 0.8781 & \textbf{0.2460} & \textbf{0.1080} & \textbf{1.6519} & \textbf{0.9909} & \textbf{1} \\
     & GEARS & 0.8637 & 0.2874 & 0.1152 & 1.7956 & 0.9347 & 2 \\
     & GenePert & \textbf{0.8845} & 0.2490 & 0.3338 & 14.1812 & 0.4982 & 3 \\
     & scGPT & 0.8568 & 0.3282 & 0.3281 & 14.0641 & 0.4197 & 4 \\
     & CPA & 0.7079 & 0.6594 & 0.2785 & 8.4113 & 0.1764 & 5 \\
    \bottomrule
  \end{tabular*}
\end{table}

\subsection{Ablations}

Table~\ref{tab:ablation} reports SciPlex3 dictionary-capacity sensitivity and removals of core components.

\begin{table}[H]
  \caption{SciPlex3 ablations. The score drop is measured from the default MRFlow score of 1.0000.}
  \label{tab:ablation}
  \centering
  \scriptsize
  \renewcommand{\arraystretch}{1.02}
  \setlength{\tabcolsep}{1pt}
  \begin{tabular*}{\linewidth}{@{\extracolsep{\fill}}lrrrrr@{}}
    \toprule
    Setting & Pearson & MSE & E-dist. & Score & Drop \\
    \midrule
    shared dictionary \(K=16\) & \textbf{0.9608} & \textbf{0.6957} & \textbf{9.9167} & \textbf{1.0000} & 0.0000 \\
    shared dictionary \(K=32\) & 0.9589 & 0.7136 & 10.0382 & 0.9544 & 0.0456 \\
    shared dictionary \(K=8\) & 0.9542 & 0.7428 & 10.2146 & 0.8749 & 0.1251 \\
    shared dictionary \(K=4\) & 0.9419 & 0.8617 & 10.7458 & 0.6226 & 0.3774 \\
    shared dictionary \(K=2\) & 0.9236 & 1.0185 & 11.3624 & 0.2929 & 0.7071 \\
    single-basis dictionary \(K=1\) & 0.9018 & 1.1403 & 11.8071 & 0.0000 & 1.0000 \\
    \midrule
    w/o entropic OT alignment & 0.9439 & 0.7002 & 13.8147 & 0.7962 & 0.2038 \\
    w/o DE weighting & 0.9135 & 0.8847 & 12.5839 & 0.7435 & 0.2565 \\
    w/o OT-guided pairing & 0.9576 & 0.7712 & 17.0063 & 0.6317 & 0.3683 \\
    w/o low-rank correction & 0.9055 & 1.1067 & 13.3440 & 0.6065 & 0.3935 \\
    w/o residual field & 0.8731 & 1.1564 & 12.7894 & 0.5760 & 0.4240 \\
    w/o KL regularization & 0.8848 & 1.2655 & 13.5225 & 0.5090 & 0.4910 \\
    w/o macro field & 0.6595 & 1.4959 & 13.4944 & 0.1651 & 0.8349 \\
    \bottomrule
  \end{tabular*}
\end{table}

Increasing the shared dictionary size from \(K=1\) to \(K=16\) improves all metrics, while \(K=32\) gives no further gain, consistent with the dictionary acting as a regularized response basis rather than as capacity alone. Removing the macro field causes the strongest failure, dropping DE20 Pearson from 0.9608 to 0.6595; removing the residual field also degrades performance, indicating that affine response modes alone do not capture nonlinear single-cell heterogeneity. Random pairing most clearly affects distributional error, increasing E-distance from 9.9167 to 17.0063. Appendix~\ref{app:add_quant} provides the corresponding degradation heatmap and dictionary-capacity curve. These ablations are conducted on SciPlex3, the largest drug perturbation benchmark in this study.

\subsection{Interpretability and dynamics}

\begin{figure}[H]
  \centering
  \includegraphics[width=0.92\linewidth]{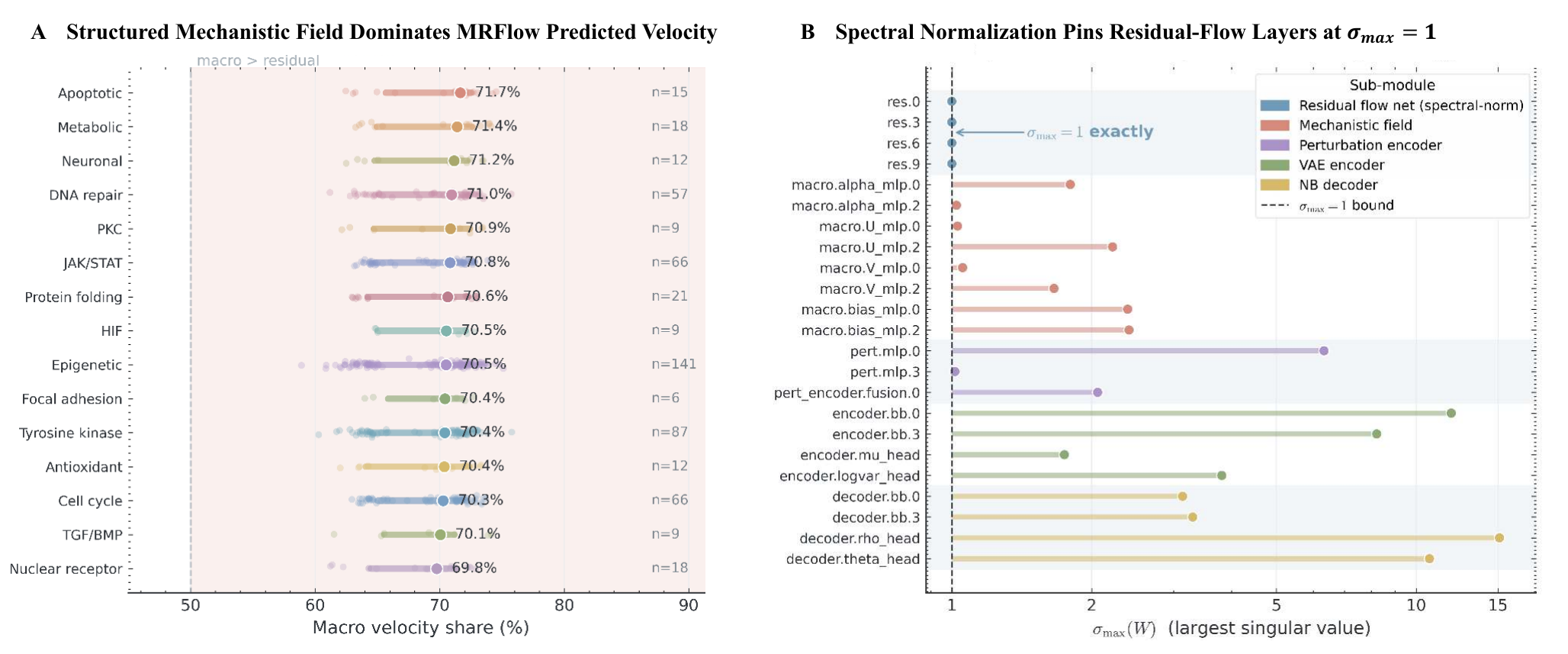}
  \caption{Mechanism-structured and residual velocity diagnostics. The macro field accounts for most predicted velocity, while spectral normalization keeps residual-flow layers controlled.}
  \label{fig:structured}
\end{figure}

\begin{figure}[H]
  \centering
  \includegraphics[width=0.95\linewidth,trim=0 12 0 10,clip]{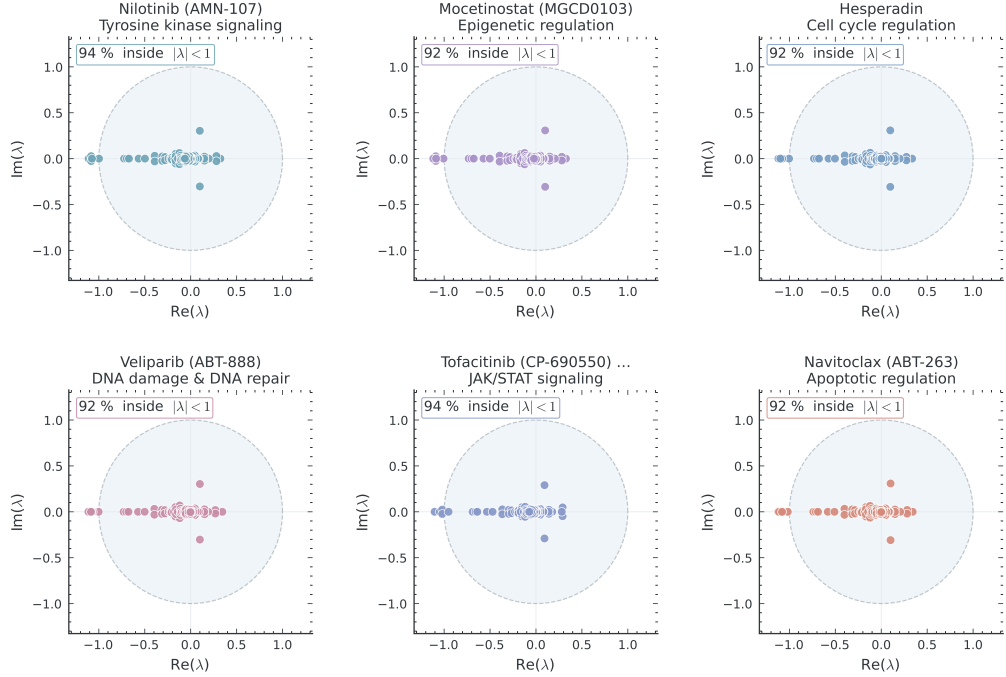}
  \caption{Eigenvalue spectra of \(W(c)\). Most modes remain inside the unit circle; outside modes mark drug-specific strong responses.}
  \label{fig:eigenvalues}
\end{figure}

The mechanism-structured residual split provides operator-level diagnostics for the learned transport. Figure~\ref{fig:structured} shows that the macro field contributes slightly more than half of the predicted velocity across major SciPlex3 drug pathway groups, with pathway-level macro shares from 51.5\% to 52.7\%; the residual branch remains substantial but not dominant. The learned affine operators can also be inspected through their spectra. Figure~\ref{fig:eigenvalues} shows that about 92\% to 94\% of eigenvalue modes satisfy \(|\lambda|<1\), while the few outside modes are better interpreted as drug-specific strong-response directions than as global instability.

\begin{figure}[H]
  \centering
  \includegraphics[width=\linewidth]{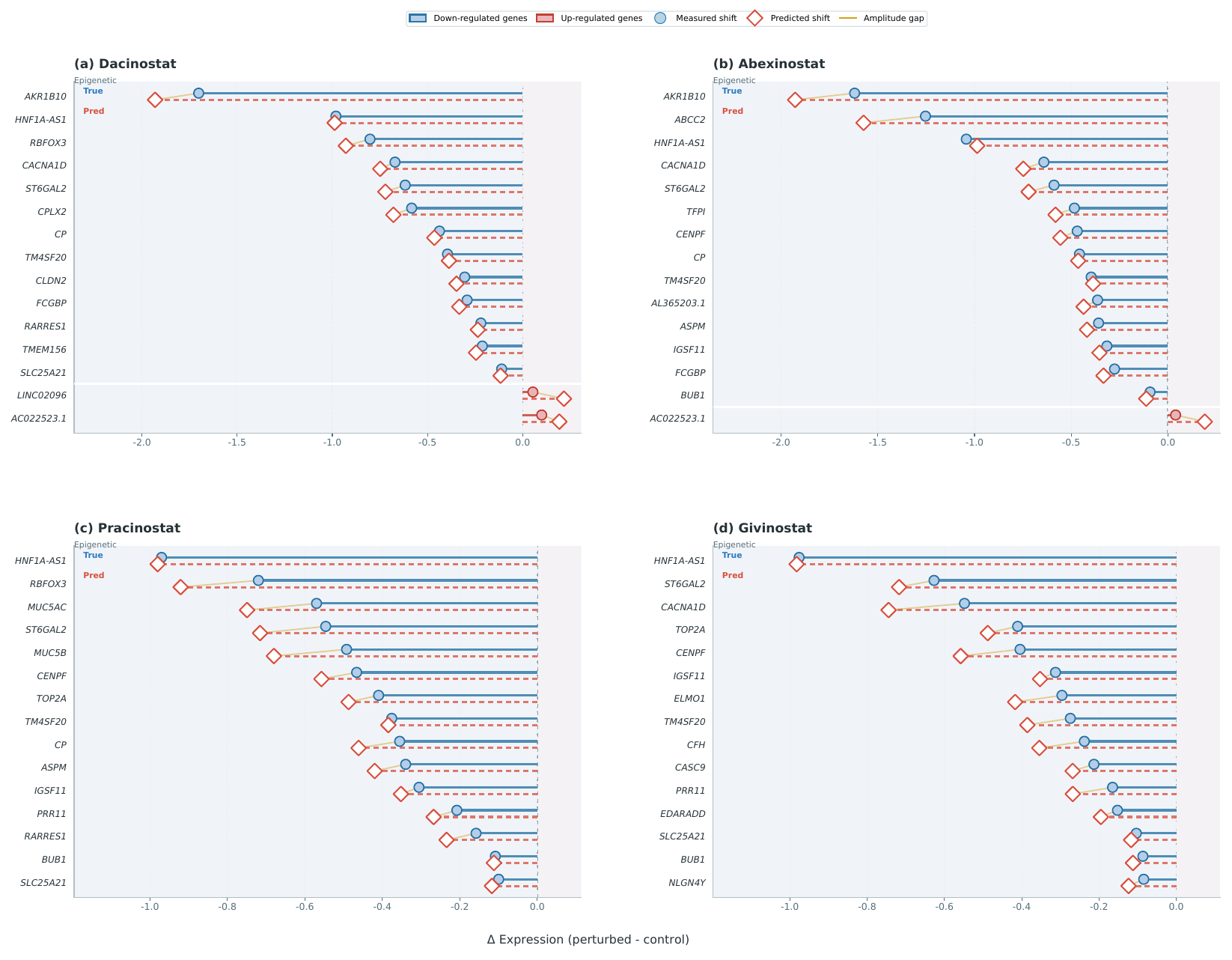}
  \caption{HDAC inhibitor response-gene shifts. MRFlow preserves selected downstream markers.}
  \label{fig:hdac}
\end{figure}

At gene level, Figure~\ref{fig:hdac} compares measured and predicted shifts for four HDAC inhibitors in A549 cells at 1000 nM. The genes are downstream response markers, not asserted direct HDAC targets; agreement therefore checks response geometry without claiming recovery of the HDAC regulatory program.

Appendices~\ref{app:table_si} and~\ref{app:matched_shift_examples} provide mechanism-structured diagnostics, provenance tables, and selected-condition examples; the main claims remain anchored in raw metrics and the shared evaluation protocol. These panels make distributional prediction tangible at the scale of individual response genes. They complement Tables~\ref{tab:drug_raw} and~\ref{tab:gene_raw} rather than replacing them: the tables quantify held-out performance across all evaluated conditions, while the panels ask whether selected signed gene shifts retain interpretable direction and relative amplitude. This distinction matters because a model can obtain a high mean-effect correlation while still placing the predicted population in an implausible region of expression space.

The visual diagnostics also clarify the role of the mechanism-structured field. The affine dictionary constrains the dominant response to reusable latent operators, and the residual branch accounts for condition- and cell-state-specific variation. The HDAC examples therefore serve as a local consistency check on the same design principle evaluated quantitatively in the ablations: structured response modes should preserve broad perturbation direction, while residual dynamics should refine heterogeneous expression shifts without turning the model into an unconstrained generator.

\section{Conclusion}

MRFlow casts virtual cell perturbation prediction as conditional latent distribution transport, combining OT-guided flow matching with a mechanism-structured affine dictionary, residual correction, and count-aware decoding. Across eight benchmarks, raw metrics and ablations support evaluating virtual cells as population generators rather than only mean-response regressors. MRFlow remains a task-level architecture trained per benchmark, but it offers a concrete route toward predictors judged by differential-expression accuracy, population distance, and response-gene consistency.

\clearpage

\bibliographystyle{unsrtnat}
\bibliography{references}

\clearpage
\appendix
\setcounter{table}{0}
\setcounter{figure}{0}
\setcounter{equation}{0}
\renewcommand{\thetable}{\thesection.\arabic{table}}
\renewcommand{\thefigure}{\thesection.\arabic{figure}}
\renewcommand{\theequation}{\thesection.\arabic{equation}}
\renewcommand{\theHtable}{\thesection.\arabic{table}}
\renewcommand{\theHfigure}{\thesection.\arabic{figure}}
\renewcommand{\theHequation}{\thesection.\arabic{equation}}
\makeatletter
\@addtoreset{table}{section}
\@addtoreset{figure}{section}
\@addtoreset{equation}{section}
\makeatother

\section{Raw Metrics Across Benchmarks}
\label{app:rawmetrics}

\begin{figure}[H]
  \centering
  \includegraphics[width=\linewidth]{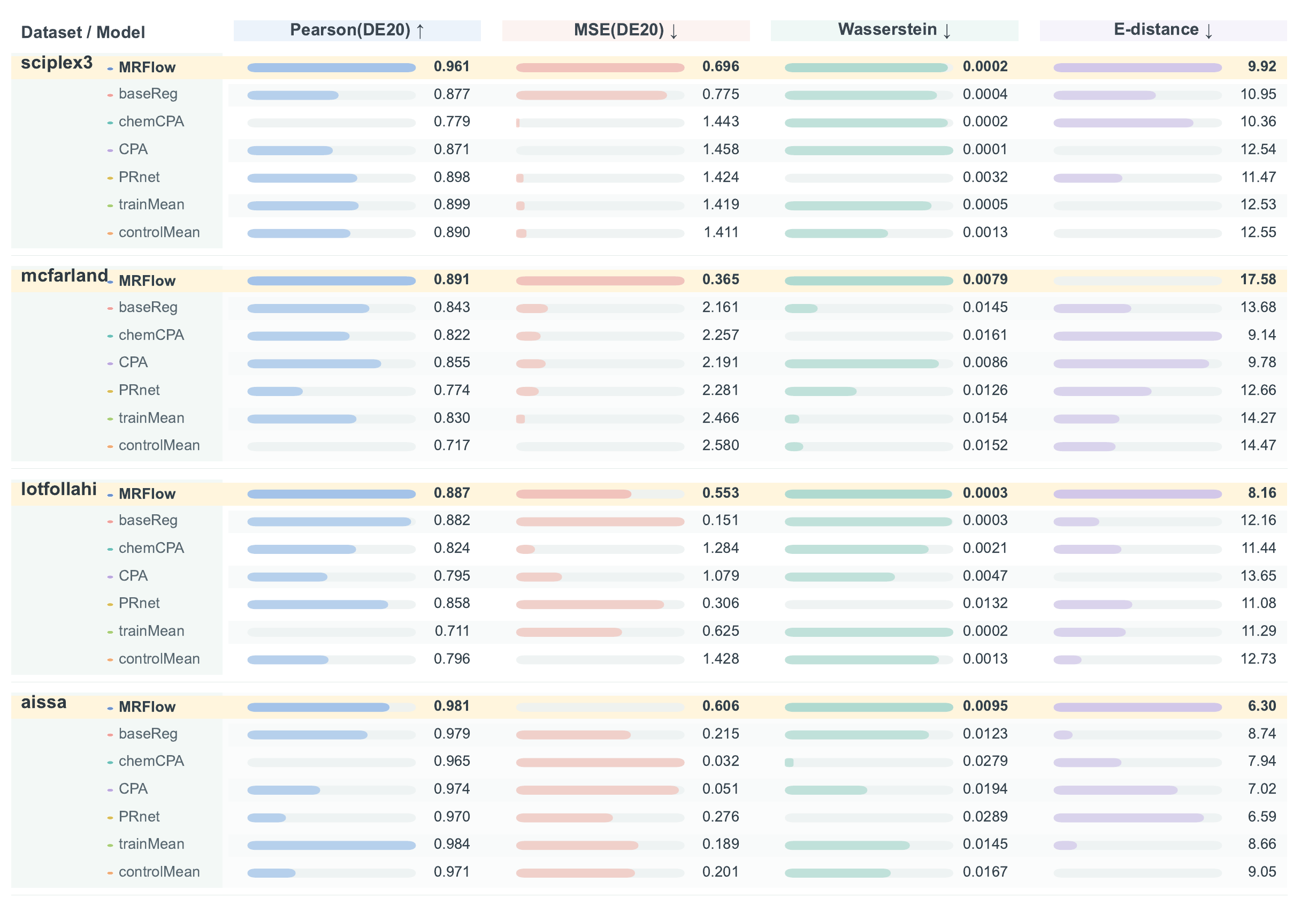}
  \caption{Drug perturbation benchmark comparison. The visualization summarizes raw-metric behavior across drug perturbation datasets using the same held-out condition evaluations reported in Table~\ref{tab:drug_raw}.}
  \label{fig:app_drug_comparison}
\end{figure}

\begin{figure}[H]
  \centering
  \includegraphics[width=\linewidth]{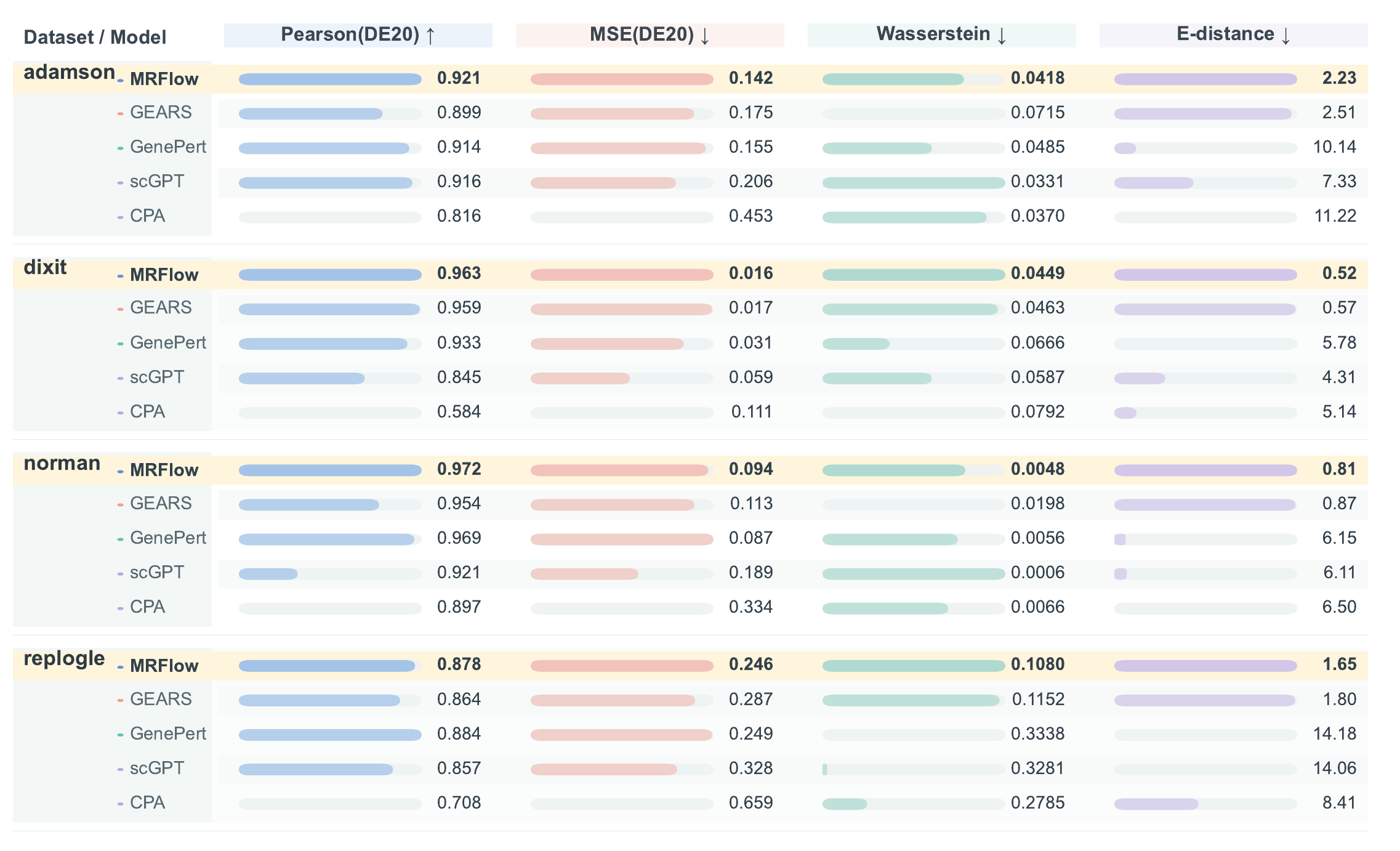}
  \caption{Genetic perturbation benchmark comparison. The visualization summarizes raw-metric behavior across genetic perturbation datasets using the same held-out condition evaluations reported in Table~\ref{tab:gene_raw}.}
  \label{fig:app_gene_comparison}
\end{figure}

The raw metrics across drug and genetic perturbation benchmarks are reported in Tables~\ref{tab:drug_raw} and~\ref{tab:gene_raw} in the main text. Figures~\ref{fig:app_drug_comparison} and~\ref{fig:app_gene_comparison} provide companion visualizations of the same benchmark comparisons, separating drug and genetic perturbation tasks. The composite score is an aggregate summary computed from these four raw metrics, with the normalization rule specified in Appendix~\ref{app:composite}.

This section makes the evidence behind the main-text ranking inspectable without relying on the composite score alone. The visualizations present the comparison from a metric-level perspective, separating mean-effect accuracy, measured by DE20 Pearson and DE20 MSE, from population-level agreement, measured by Wasserstein distance and E-distance. This distinction is central to MRFlow: the model is designed as a distributional transport method, and its advantage should appear not only in average expression shifts but also in the geometry of predicted cell populations.

The four metrics are non-redundant by design. DE20 Pearson asks whether the strongest differentially expressed genes move in the correct relative direction, while DE20 MSE penalizes the absolute size of those shifts. Wasserstein distance and E-distance instead compare empirical distributions of predicted and observed cell populations. A method can therefore score well on the mean response while producing a poor population distribution, or it can improve distributional agreement at the cost of a small loss in mean-effect correlation. The appendix figures make these trade-offs visible because a single composite number would hide which biological question each method answers well.

This metric separation is also needed for comparing drug and genetic perturbation tasks. Drug screens contain dose and cell-context variation, so strong average-response prediction can coexist with heterogeneous single-cell effects. Genetic screens often include sharper transcriptional shifts, but the source and target populations remain unpaired. MRFlow addresses both regimes through the same conditional transport architecture, and the raw-metric view directly tests whether the architecture behaves consistently across modalities.

The drug benchmark visualization supports the main-text claim that MRFlow gains are not an artifact of normalization. On SciPlex3, MRFlow gives the strongest DE20 Pearson, DE20 MSE, and E-distance among the compared methods, showing that the latent transport objective improves both mean-effect recovery and response-distribution agreement in the largest drug screen. McFarland and Aissa show more metric-specific trade-offs. These trade-offs bound the claim: MRFlow still ranks best by the aggregate comparison, while the raw view makes clear where individual baselines remain competitive on a single metric. The drug results therefore support balanced behavior across complementary response criteria, not universal dominance on every scalar score.

The Aissa benchmark is interpreted with additional caution because the processed evaluation contains only three non-control drug conditions, as documented in Appendix~\ref{app:datasets}. With so few held-out conditions, a single condition can shift an aggregate metric more strongly than in SciPlex3 or Lotfollahi. Aissa therefore functions as a small-regime stress case within the broader drug-screen analysis. Reporting raw values alongside the composite score keeps this limitation visible.

The genetic benchmark visualization gives a sharper distributional picture. MRFlow achieves the best composite score on Adamson, Dixit, Norman, and Replogle while preserving competitive DE20 Pearson. The Replogle comparison is particularly informative: GenePert attains a slightly higher Pearson score, yet MRFlow has much lower Wasserstein distance and E-distance, indicating that a model can match average differential expression while still missing the population-level response distribution. This pattern mirrors the motivation in the main text, where virtual cell prediction is framed as conditional distribution transport rather than conditional mean regression alone.

The genetic results clarify the role of distributional evaluation. On Replogle, the benchmark contains a large CRISPRi perturbation landscape with internal non-targeting control structure. A model that predicts the correct average direction for response genes may still place predicted cells in the wrong region of the response distribution. The lower distributional distances of MRFlow in this setting are therefore closely tied to the method design: the flow is trained to move empirical control populations toward empirical perturbed populations, rather than to output only a condition-level expression vector.

\section{Composite Score Definition}
\label{app:composite}

For each dataset and metric, normalization is computed independently across the compared methods on that dataset. For Pearson correlation, where larger values are better, the normalized score for a method value \(x\) is
\begin{equation}
\operatorname{Norm}(x)=\frac{x-x_{\min}}{x_{\max}-x_{\min}} .
\end{equation}
For DE20 MSE, Wasserstein distance, and E-distance, where smaller values are better, the normalized score is
\begin{equation}
\operatorname{Norm}(x)=\frac{x_{\max}-x}{x_{\max}-x_{\min}} .
\end{equation}
The composite score is the equal-weight average of the four normalized metrics,
\begin{equation}
\operatorname{Composite}
=
\frac{1}{4}
\left(
\operatorname{Norm}_{Pearson}
+\operatorname{Norm}_{MSE}
+\operatorname{Norm}_{Wasserstein}
+\operatorname{Norm}_{E}
\right).
\end{equation}
The composite score is used only as an aggregate summary; raw metrics are reported in Tables~\ref{tab:drug_raw} and~\ref{tab:gene_raw}.

Per-dataset normalization is used because the numerical scale of each metric differs across datasets. A Wasserstein distance from one dataset is not directly comparable to a Wasserstein distance from another dataset when the cell context, gene vocabulary, perturbation strength, and number of conditions differ. Normalizing within each dataset and metric makes the composite score a within-benchmark summary of relative performance, not an absolute biological distance. This design prevents a large-scale dataset with numerically larger distances from dominating the aggregate comparison.

Equal weighting avoids privileging mean-response accuracy over distributional agreement. Pearson and MSE summarize differential-expression recovery, while Wasserstein distance and E-distance summarize population-level agreement. The composite therefore assigns half of its weight to mean-effect behavior and half to distributional behavior. This choice matches the paper's formulation of virtual cell prediction as population transport, while retaining the familiar DE20 metrics used in perturbation-prediction benchmarks.

The composite score is defined as an aggregate summary of the raw metrics. It provides a compact ranking because each benchmark contains multiple methods and four non-commensurate metrics, but it is not a substitute for the underlying measurements. For this reason, the main text reports Tables~\ref{tab:drug_raw} and~\ref{tab:gene_raw}, Appendix~\ref{app:rawmetrics} visualizes the raw comparisons, and Appendix~\ref{app:add_quant} summarizes raw-metric ranks without reusing the composite normalization.

\section{Dataset and Split Details}
\label{app:datasets}

Table~\ref{tab:dataset_details} summarizes the datasets used in the benchmark. Each benchmark uses a condition-level held-out protocol with test ratio \(0.2\) for datasets evaluated under random condition splits. A condition is defined by the perturbation identity together with the dose and cell context whenever those annotations are present. The split is therefore a held-out perturbation-condition split rather than an unseen-drug-identity or unseen-gene-identity split.

The eight benchmarks stress different parts of the virtual-cell prediction problem. Norman, Adamson, Dixit, and Replogle evaluate genetic perturbations across CRISPRa, CRISPRi, and CRISPR knockout designs. SciPlex3, McFarland, Aissa, and Lotfollahi evaluate chemical perturbations across multi-cell-line screens, dose variation, resistance-associated drug response, and drug-combination settings. This mixture tests whether a model can maintain the same response-distribution principle across perturbation technologies and cell contexts.

The held-out unit is a perturbation condition rather than an individual cell. This matters because cells sampled from the same condition are exchangeable observations of one experimental response distribution. Randomly splitting cells within a condition would test interpolation among replicate cells, which is easier and less relevant for AIVC use cases. The condition-level split instead asks whether a model can predict the response distribution for conditions that were not observed during training under the benchmark split.

Several dataset-specific details affect interpretation. Replogle uses non-targeting sgRNA as an internal control rather than a separate control-cell group, so control information is defined by the screen design. Aissa has only three non-control drug conditions in the processed benchmark, which makes any aggregate statistic more sensitive to individual condition behavior. SciPlex3 and McFarland contain explicit cell-context variation, so the condition includes context in addition to perturbation identity. These details are not exceptions to the evaluation protocol; they define what a held-out condition means for each dataset.

\begin{table}[H]
  \caption{Dataset and held-out split details. Replogle uses non-targeting sgRNA as an internal control rather than a separate control-cell group.}
  \label{tab:dataset_details}
  \centering
  \scriptsize
  \renewcommand{\arraystretch}{1.08}
  \setlength{\tabcolsep}{3pt}
  \resizebox{\linewidth}{!}{%
  \begin{tabular}{lllrrrrrp{0.30\linewidth}}
    \toprule
    Dataset & Type & Cell context & Cells & HVG & Control & Perturbed & Conditions & Held-out protocol \\
    \midrule
    Norman & Gene & K562 & 111255 & 2053 & 11835 & 99420 & 236 & Condition-level held-out split with a shared evaluation protocol across MRFlow and all compared methods. \\
    Replogle & Gene & RPE1 & 162733 & 2431 & 0 & 162733 & 1544 & Condition-level held-out split; non-targeting sgRNA is used as internal control. \\
    Adamson & Gene & K562 & 68603 & 2077 & 24263 & 44340 & 86 & Condition-level held-out split with a shared evaluation protocol across MRFlow and all compared methods. \\
    Dixit & Gene & K562 & 44735 & 2013 & 12042 & 32693 & 19 & Condition-level held-out split with a shared evaluation protocol across MRFlow and all compared methods. \\
    SciPlex3 & Drug & MCF7/A549/K562 & 799317 & 2000 & 17578 & 781739 & 189 & Condition-level held-out split over drug, dose, and cell-context conditions. \\
    McFarland & Drug & 169 cancer cell lines & 130648 & 2000 & 25851 & 104797 & 13 & Condition-level held-out split over drug and cell-context conditions. \\
    Aissa & Drug & PC9 & 119071 & 2000 & 29983 & 89088 & 3 & Held-out condition evaluation with only three non-control drug conditions, making aggregate statistics less stable. \\
    Lotfollahi & Drug & A549 & 63430 & 2000 & 1451 & 61979 & 31 & Condition-level held-out split over single-drug and combination conditions. \\
    \bottomrule
  \end{tabular}}
\end{table}

The shared preprocessing and split protocol is part of the evidence chain. Every method is evaluated on the same processed expression matrix and the same held-out conditions within a dataset, so differences in Tables~\ref{tab:drug_raw} and~\ref{tab:gene_raw} reflect model behavior under matched inputs rather than data leakage, alternative gene vocabularies, or different evaluation partitions. The table also explains why the paper avoids stronger claims such as strictly unseen drug identity or strictly unseen gene identity. The benchmark is a held-out condition evaluation, and a condition may combine perturbation identity, dose, and cell context.

\section{OT-Guided Pairing Implementation}
\label{app:pairing}

MRFlow uses low-cost latent pairing to construct flow-matching supervision from unpaired perturbation measurements. At each training step, the sampler chooses a perturbation condition \(c\), draws \(N\) control cells and \(N\) perturbed cells under that condition, and encodes them as \(Z_0=\{z_i^0\}_{i=1}^{N}\) and \(Z_1=\{z_j^1\}_{j=1}^{N}\). The pairing cost is computed from detached latent states,
\begin{equation}
C_{ij}=\left\|\operatorname{stopgrad}(z_i^0)-\operatorname{stopgrad}(z_j^1)\right\|_2^2 .
\end{equation}
The default implementation uses greedy nearest-neighbor low-cost pairing,
\begin{equation}
\tau(i)=\arg\min_j C_{ij},
\qquad
z_{1,i}^{\pi}=z_{\tau(i)}^1 .
\end{equation}
The pairing is recomputed for every minibatch and is not itself differentiated. Gradients are taken through the flow-matching loss and the differentiable encoder, velocity field, and decoder computations. The random-pairing ablation in Table~\ref{tab:ablation} replaces this low-cost pairing with random endpoint matching. The reported implementation uses greedy low-cost latent pairing as the default coupling rule; all results in the main tables and ablations are based on this pairing protocol.

The pairing defines stable training supervision; it does not assert that a particular control cell is the biological ancestor of a particular perturbed cell. Perturbation screens are unpaired because the same cell cannot be observed before and after intervention. A random pairing would create many long and crossing interpolation paths in latent space, forcing the vector field to explain artificial transport directions. Greedy low-cost pairing reduces this avoidable noise by matching each encoded control state to a nearby perturbed endpoint within the same condition, which gives the flow-matching objective a more coherent local velocity target.

The stop-gradient operation is part of the supervision design. If gradients passed through the pairing cost, the encoder could reduce the loss by changing the geometry used to choose pairs rather than by learning a predictive transport field. Detaching the latent states for the cost matrix keeps the assignment step as a minibatch-level supervision construction. The differentiable learning signal then comes from the velocity regression, reconstruction, variational regularization, and endpoint-alignment terms described in Appendix~\ref{app:objective}.

\begin{table}[H]
  \caption{Algorithm: low-cost latent pairing for flow matching.}
  \label{tab:pairing_algorithm}
  \centering
  \small
  \renewcommand{\arraystretch}{1.12}
  \setlength{\tabcolsep}{3pt}
  \begin{tabular*}{\linewidth}{@{\extracolsep{\fill}}p{0.22\linewidth}p{0.70\linewidth}@{}}
    \toprule
    Operation & Description \\
    \midrule
    Condition sampling & Select a perturbation condition \(c\) from the training conditions. \\
    Cell sampling & Draw \(N\) control cells and \(N\) perturbed cells under condition \(c\). \\
    Encoding & Compute latent states \(z_i^0\sim q_\phi(z\mid x_i^0)\) and \(z_j^1\sim q_\phi(z\mid x_j^c)\). \\
    Cost construction & Compute \(C_{ij}=\|\operatorname{stopgrad}(z_i^0)-\operatorname{stopgrad}(z_j^1)\|_2^2\). \\
    Pairing & Set \(\tau(i)=\arg\min_j C_{ij}\) and pair \(z_i^0\) with \(z_{1,i}^{\pi}=z_{\tau(i)}^1\). \\
    Flow supervision & Sample \(t\sim\mathcal{U}(0,1)\), form \(z_t=(1-t)z_i^0+t z_{1,i}^{\pi}\), and use \(z_{1,i}^{\pi}-z_i^0\) as the velocity target. \\
    \bottomrule
  \end{tabular*}
\end{table}

The auxiliary OT alignment term in Equation~\ref{eq:objective} is implemented as an entropic OT alignment cost between predicted endpoints and paired targets, distinct from a debiased Sinkhorn divergence.

This implementation choice also explains the ablation behavior in Table~\ref{tab:ablation} and Figure~\ref{fig:app_ablation_degradation}. Replacing low-cost pairing with random pairing leaves a relatively high DE20 Pearson but substantially worsens E-distance, showing that mean differential-expression recovery can survive noisy pairing while population-level geometry suffers. The result supports the interpretation that OT-guided pairing mainly stabilizes the distributional side of the prediction task.

\section{Training Objective and Stabilization}
\label{app:objective}

The training objective combines a transport loss with count modeling and regularization because these terms address different failure modes. The flow-matching term trains the conditional vector field to move latent control states toward paired perturbed endpoints. The negative binomial reconstruction term keeps latent states tied to count-aware expression distributions. The KL term regularizes the variational cell-state space, and the structured-field penalties prevent the residual and affine components from absorbing variation through unconstrained magnitude. The auxiliary entropic OT term adds endpoint-level alignment between predicted and paired target populations.

The flow-matching loss is
\begin{equation}
\mathcal{L}_{FM}
=
\mathbb{E}
\left[
\left\|v_\Theta(z_t,t,c)-(z_1^\pi-z_0)\right\|_2^2
\right].
\end{equation}
The reconstruction term is the negative binomial negative log likelihood for control and perturbed cells. For count \(y_{ig}\), mean \(\mu_{ig}=\ell_i\rho_{\theta,g}(z_i)\), and dispersion \(\theta_{\theta,g}(z_i)\), the likelihood is
\begin{equation}
p(y_{ig}\mid z_i,\ell_i)
=
\frac{\Gamma(y_{ig}+\theta_{\theta,g})}
{\Gamma(\theta_{\theta,g})\Gamma(y_{ig}+1)}
\left(
\frac{\theta_{\theta,g}}{\theta_{\theta,g}+\mu_{ig}}
\right)^{\theta_{\theta,g}}
\left(
\frac{\mu_{ig}}{\theta_{\theta,g}+\mu_{ig}}
\right)^{y_{ig}} .
\end{equation}
The KL regularizer is
\begin{equation}
\mathcal{L}_{KL}
=
\frac{1}{2}
\mathbb{E}
\sum_{m=1}^{d}
\left(
\mu_m^2+\sigma_m^2-\log\sigma_m^2-1
\right).
\end{equation}
The residual and affine penalties are
\begin{equation}
\mathcal{L}_{res}=\mathbb{E}\|v_{\mathrm{res}}(z_t,t,c)\|_2^2,
\qquad
\mathcal{L}_{W}=\mathbb{E}_{c}\|W(c)\|_F^2 .
\end{equation}
The auxiliary alignment cost compares predicted endpoints with paired perturbed endpoints through entropic OT,
\begin{equation}
\mathcal{L}_{OT}^{aux}
=
OT_{\varepsilon}
\left(
\{\hat{z}_{1,i}\}_{i=1}^{N},
\{z_{1,i}^{\pi}\}_{i=1}^{N}
\right),
\qquad
\hat{z}_{1,i}=z_{t,i}+(1-t)v_\Theta(z_{t,i},t,c).
\end{equation}

Differentially expressed genes are upweighted in the reconstruction term through a condition-specific gene weight
\begin{equation}
w_g(c)=1+\eta_{DE}\mathbf{1}\left[g\in\mathcal{D}_{K}(c)\right],
\end{equation}
where \(\mathcal{D}_{K}(c)\) denotes the selected differentially expressed genes for condition \(c\). This weighting is a training stabilization strategy rather than a separate model component. The implementation uses KL warmup and free-bits style stabilization, replacing the raw KL contribution with a warmup-weighted term and preventing very small per-dimension KL values from dominating posterior collapse behavior. Condition-balanced sampling is used for imbalanced datasets, assigning higher sampling probability to conditions with fewer cells. These stabilization details are not used to introduce additional evaluation claims.

The objective should be interpreted as a practical training principle rather than a theorem that each term uniquely identifies a biological mechanism. The ablations show that removing terms changes empirical behavior on SciPlex3, but they do not prove causal necessity in every dataset. The macro field, residual field, low-rank correction, KL regularization, entropic OT alignment, DE weighting, and OT-guided pairing operate together. Their roles are separable enough to produce distinct degradation patterns, yet the final model is trained as a coupled system.

DE gene weighting is included because benchmark metrics emphasize response genes. Without this weighting, the reconstruction loss can be dominated by genes whose expression is stable under a condition, reducing pressure to model perturbation-induced shifts. The weighting does not change the evaluation target and does not insert test labels into inference; it only adjusts the training emphasis for genes selected within the training procedure for each condition.

Condition-balanced sampling addresses another practical issue in perturbation atlases. Some perturbation conditions contain many more cells than others. Sampling conditions more evenly prevents large conditions from dominating the velocity field and decoder updates. For distributional prediction, the model must learn response geometry across many conditions rather than optimizing primarily for the most frequently sampled ones.

\section{Baseline and Preprocessing Details}
\label{app:baseline}

All methods are evaluated on the same held-out perturbation conditions for each dataset, and metrics are computed using one shared evaluation script. Preprocessing follows the dataset-specific HVG selections summarized in Appendix~\ref{app:datasets}; perturbation genes are kept in the modeling vocabulary under the benchmark vocabulary construction. Expression metrics are computed in the same expression space for all methods within a dataset, while the negative binomial decoder is trained against count-scale targets for datasets with count matrices.

Table~\ref{tab:baseline_fairness} specifies the role and implementation source for each baseline. CPA, chemCPA, PRnet, GEARS, GenePert, and scGPT are run from official open-source implementations. controlMean, trainMean, and baseReg are deterministic benchmark baselines evaluated from the same preprocessed matrices. Within each dataset, every method receives the same train and test perturbation-condition split, the same processed expression matrix, the same HVG and benchmark gene vocabulary, and the same metric computation pipeline.

The baselines cover three levels of difficulty. controlMean and trainMean test whether a method beats simple population summaries. baseReg tests a deliberately simple supervised regression rule, motivated by recent perturbation-prediction studies showing that simple baselines can be highly competitive. CPA, chemCPA, PRnet, GEARS, GenePert, and scGPT represent learned perturbation models that differ in condition encoding, graph use, chemical structure handling, and pretraining. Comparing against all three groups reduces the chance that MRFlow is only outperforming weak baselines or competing within a single model family.

The fairness principle is input alignment. A baseline does not receive a different held-out split, a different expression matrix, or a different metric definition. The shared setup is especially important for distributional metrics because Wasserstein distance and E-distance can change with preprocessing choices, gene vocabulary, and cell subsampling. Keeping these inputs fixed makes the raw-metric comparisons in Tables~\ref{tab:drug_raw} and~\ref{tab:gene_raw} attributable to the predictive models.

All datasets and learned baseline implementations used in this study originate from public papers, public data portals, or official open-source releases associated with the cited work. The final reproduction package will preserve attribution to the original creators and will follow the corresponding licenses, access requirements, and redistribution terms for each dataset and software component.

\begin{table}[H]
  \caption{Baseline implementation and evaluation alignment.}
  \label{tab:baseline_fairness}
  \centering
  \scriptsize
  \setlength{\tabcolsep}{3pt}
  \renewcommand{\arraystretch}{1.08}
  \begin{tabular*}{\linewidth}{@{\extracolsep{\fill}}lll@{}}
    \toprule
    Baseline & Role & Implementation and shared evaluation inputs \\
    \midrule
    controlMean & Control-population mean predictor & Deterministic benchmark rule; same split, matrix, vocabulary, and metrics. \\
    trainMean & Training-condition mean predictor & Deterministic benchmark rule; same split, matrix, vocabulary, and metrics. \\
    baseReg & Regression baseline & Deterministic benchmark regression; same split, matrix, vocabulary, and metrics. \\
    CPA & Conditional perturbation autoencoder & Official open-source implementation; same split, matrix, vocabulary, and metrics. \\
    chemCPA & Structure-aware chemical CPA & Official open-source implementation; same split, matrix, vocabulary, and metrics. \\
    PRnet & Drug perturbation generative model & Official open-source implementation; same split, matrix, vocabulary, and metrics. \\
    GEARS & Graph gene perturbation model & Official open-source implementation; same split, matrix, vocabulary, and metrics. \\
    GenePert & Gene-embedding linear model & Official open-source implementation; same split, matrix, vocabulary, and metrics. \\
    scGPT & Single-cell foundation model & Official open-source implementation; same split, matrix, vocabulary, and metrics. \\
    \bottomrule
  \end{tabular*}
\end{table}

The benchmark does not report confidence intervals across random seeds. The results are therefore a single-run comparison under a fixed evaluation protocol. This limitation does not affect the consistency of the shared split and metric computation, but seed-level uncertainty remains outside the claims made in this manuscript.

\section{Compute Resources and Broader Impact}
\label{app:compute_impact}

All MRFlow experiments were run on a workstation with one NVIDIA RTX 5090 GPU with 32 GB memory and 16 vCPUs from an Intel(R) Xeon(R) Platinum 8352V CPU at 2.10 GHz. The software environment used Ubuntu 22.04, Python 3.12, PyTorch 2.8.0, and CUDA 12.8. Training time varied with dataset size and condition complexity, ranging from approximately 3 to 8 hours per dataset. The largest screens require more time because condition-balanced sampling, latent pairing, endpoint alignment, and count reconstruction are computed over many perturbation conditions.

MRFlow supports perturbation prioritization, drug-response analysis, and virtual-cell modeling by predicting response distributions before costly wet-lab validation. This can reduce experimental search space and help identify perturbation conditions that merit closer biological inspection. The main risk is over-interpretation: in-silico predictions may reflect biases in the training atlas, preprocessing choices, or benchmark splits, and they are not clinical or causal evidence without experimental validation. The paper therefore reports raw metrics, dataset details, baseline alignment, and qualitative response-gene panels to make model behavior inspectable, without presenting MRFlow as a replacement for biological experiments.

\section{Additional Quantitative Evidence}
\label{app:add_quant}

\begin{figure}[H]
  \centering
  \includegraphics[width=\linewidth]{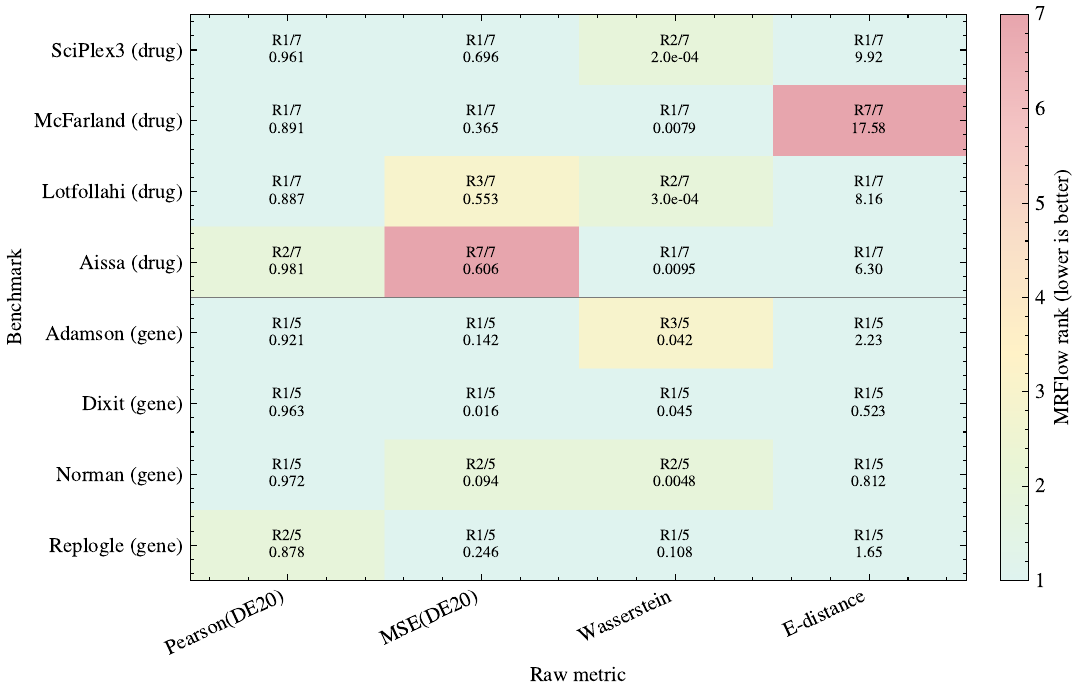}
  \caption{Raw-metric rank of MRFlow across benchmarks. Each cell reports the MRFlow rank and raw value for one dataset and one metric among the compared methods. Lower ranks are better; Pearson is ranked in descending order, and MSE, Wasserstein distance, and E-distance are ranked in ascending order.}
  \label{fig:app_raw_metric_rank}
\end{figure}

\begin{figure}[H]
  \centering
  \includegraphics[width=0.74\linewidth]{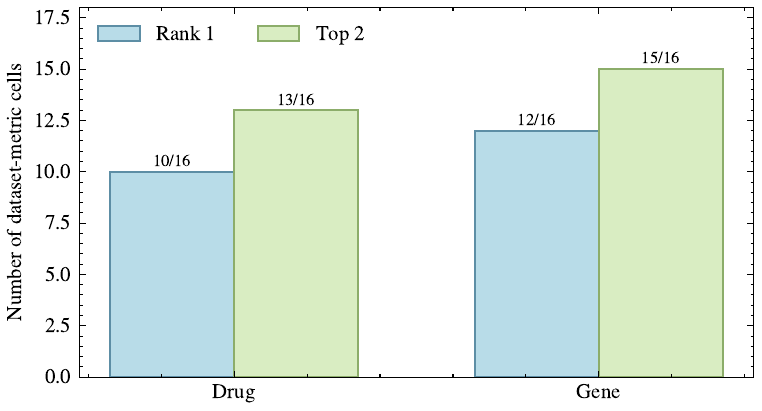}
  \caption{Raw-metric rank summary without composite scoring. Across 16 drug dataset-metric cells, MRFlow ranks first in 10 cells and within the top two in 13 cells, with mean rank 2.0625. Across 16 genetic dataset-metric cells, MRFlow ranks first in 12 cells and within the top two in 15 cells, with mean rank 1.3125.}
  \label{fig:app_raw_rank_summary}
\end{figure}

Because composite scores can obscure metric-specific trade-offs, Figures~\ref{fig:app_raw_metric_rank} and~\ref{fig:app_raw_rank_summary} summarize MRFlow ranks directly over raw metric cells. These summaries use the same values reported in Tables~\ref{tab:drug_raw} and~\ref{tab:gene_raw}; they are rank diagnostics rather than statistical significance tests. Together with Appendix~\ref{app:rawmetrics}, they address a common concern for multi-metric perturbation prediction: whether a method wins because a normalization scheme favors one metric or because it remains competitive across the raw measurements that define the task.

Raw-metric ranks are a stricter sanity check than the composite score in one respect: they ignore the magnitude of each advantage and ask only how often MRFlow is near the best method for each dataset-metric cell. A method that wins the composite through one very strong metric and several weak metrics would be exposed by this view. Conversely, a model with consistently high ranks across Pearson, MSE, Wasserstein distance, and E-distance has evidence for balanced behavior even when another method wins an isolated metric.

The rank summaries sharpen the interpretation of the main-text tables. In drug perturbation tasks, MRFlow ranks first in 10 of 16 dataset-metric cells and remains within the top two in 13 of 16 cells; in genetic perturbation tasks, it ranks first in 12 of 16 cells and within the top two in 15 of 16 cells. These counts show that MRFlow performance is not driven by one favorable dataset or one favorable metric. At the same time, the non-perfect ranks keep the claim bounded: MRFlow offers the most consistent cross-metric profile in this benchmark, not a universal win on every individual number.

The rank heatmap also makes the modality difference transparent. Drug datasets show more heterogeneous ranks, consistent with dose, cell-line, and compound-specific variation. Genetic datasets show a tighter rank profile, especially on distributional metrics. This difference does not imply that one modality is easier in general; it reflects the particular benchmark composition and the compared model set. The conclusion is deliberately narrow: under the shared evaluation protocol, MRFlow maintains strong raw-metric ranks across both modalities.

\begin{figure}[H]
  \centering
  \includegraphics[width=\linewidth]{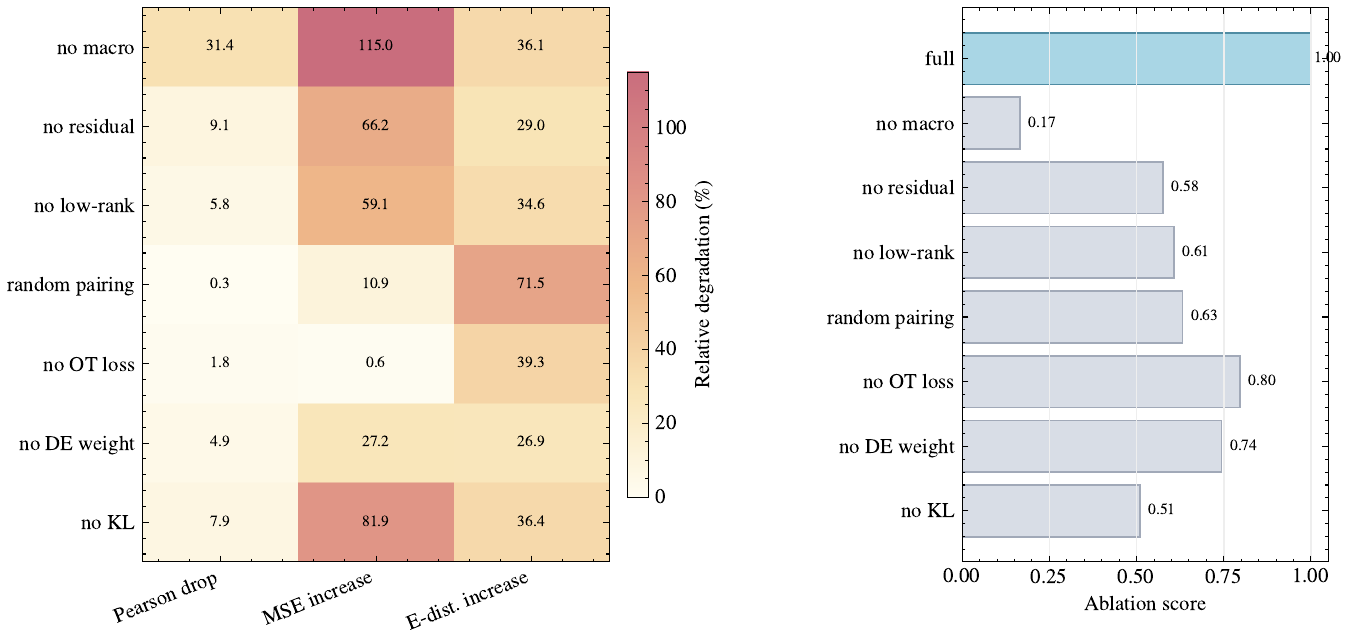}
  \caption{Component ablation on SciPlex3. The heatmap reports relative degradation from full MRFlow, and the bar panel reports the ablation score. Removing the macro field mainly damages DE20 correlation, while replacing low-cost latent pairing with random pairing mainly increases distributional error.}
  \label{fig:app_ablation_degradation}
\end{figure}

\begin{figure}[H]
  \centering
  \includegraphics[width=\linewidth]{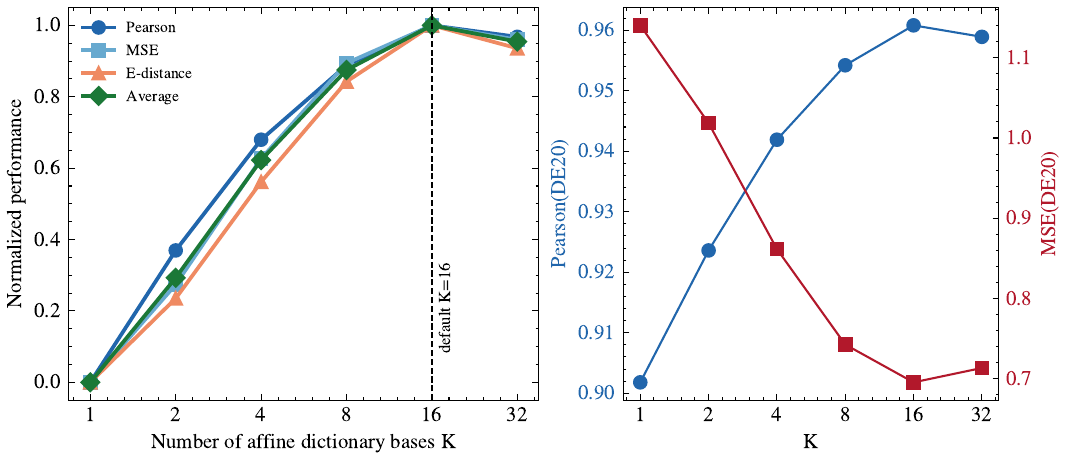}
  \caption{Sensitivity to the number of shared affine dictionary bases. Increasing \(K\) from 1 to 16 improves the SciPlex3 metrics, while \(K=32\) gives no further gain, consistent with an empirical trade-off between expressivity and regularization rather than a count of biological mechanisms.}
  \label{fig:app_k_sweep}
\end{figure}

The ablation visualizations in Figures~\ref{fig:app_ablation_degradation} and~\ref{fig:app_k_sweep} expand Table~\ref{tab:ablation}. They separate relative degradation by raw metric and show how the shared affine dictionary capacity changes mean-effect and distributional behavior on SciPlex3. These plots connect the empirical advantage back to the architecture, separating structured response modeling from a generic increase in model size.

The component ablation separates failure modes. Removing the macro field causes the largest loss in DE20 correlation, which supports the role of the mechanism-structured affine dictionary as the dominant response operator. Removing the residual field, low-rank correction, KL regularization, DE weighting, or entropic OT alignment degrades different parts of the metric profile, indicating that these terms are not interchangeable decorations around the same predictor. The random-pairing ablation is especially aligned with the distributional claim: Pearson remains relatively high, but E-distance becomes much worse, showing why low-cost latent pairing matters for cell-population agreement even when mean response correlation looks acceptable.

This ablation pattern links architectural components to measurable behavior. The macro field affects the large-scale direction of perturbation response, the residual field captures variation not explained by the affine operator, and low-cost pairing mainly protects population geometry. Different removals damage different metrics, as expected for a model whose components have distinct modeling roles.

The dictionary capacity sweep clarifies why the shared affine dictionary is a structural choice rather than a generic capacity increase. Moving from \(K=1\) to \(K=16\) improves Pearson, MSE, and E-distance, consistent with the need for multiple reusable latent response operators across perturbation conditions. Increasing to \(K=32\) does not improve the result, which suggests that too many bases can weaken the regularizing effect of the dictionary. This supports the main-text interpretation that MRFlow benefits from a controlled response basis with condition-specific recombination, not from an unconstrained velocity field.

The \(K\)-sweep is a capacity sensitivity analysis for the shared mechanism-structured affine dictionary. \(K=1\) corresponds to a single-basis dictionary and is too restrictive for the diversity of SciPlex3 perturbation responses. Intermediate values allow the model to recombine several latent response operators, and \(K=16\) gives the best reported balance in this experiment. The lack of improvement at \(K=32\) weakens a simple parameter-count explanation for the result. The evidence supports structured expressivity with regularization, not the claim that more bases always give a better virtual cell model.

\section{Supplementary Tables for Mechanism-Structured Diagnostics and Example Provenance}
\label{app:table_si}

The supplementary tables in this section make the internal evidence behind the mechanism-structured design auditable. They address a different question from the benchmark tables: rather than asking whether MRFlow predicts held-out perturbation responses better than baselines, they ask whether the learned condition-dependent operators organize perturbations in a way that is consistent with external pathway annotations, whether the macro and residual fields have separable quantitative roles, and whether the selected response-gene panels are traceable to an explicit candidate pool. These analyses support a mechanism-informed inductive bias while keeping the interpretation at the level of pathway-consistent operator organization.

The alpha-space diagnostics test the central structural hypothesis of MRFlow. For a condition \(c\), the shared affine dictionary is activated by weights \(\alpha(c)\), so the geometry of these weights reflects how the model recombines latent response operators across perturbations. If the dictionary only acted as unconstrained extra capacity, pathway labels would not be expected to concentrate in \(\alpha(c)\)-space. Tables~\ref{tab:si_alpha_summary}, \ref{tab:si_alpha_class}, and~\ref{tab:si_knn} show the opposite pattern at the diagnostic level: annotated same-class perturbations have higher alpha similarity than different-class perturbations, and nearest-neighbor enrichment exceeds permutation nulls. The level-2 pathway result is stronger than the level-1 result, which is consistent with the idea that operator reuse is more visible when broad pathway labels are refined into more specific MoA classes.

\begin{table}[H]
  \caption{Alpha-space organization by annotated perturbation class. Similarities are cosine similarities between dictionary activations \(\alpha(c)\).}
  \label{tab:si_alpha_summary}
  \centering
  \footnotesize
  \setlength{\tabcolsep}{2pt}
  \renewcommand{\arraystretch}{1.10}
  \begin{tabular*}{\linewidth}{@{\extracolsep{\fill}}lrrrrrrrr@{}}
    \toprule
    Label level & Cond. & Class & Within & Between & Difference & MW \(p\) & Perm. \(p\) & 1NN enrich. \\
    \midrule
    Pathway level 1 & 120 & 8 & 0.0332 & 0.0067 & 0.0265 & 0.0499 & 0.0740 & 2.4674 \\
    Pathway level 2 & 93 & 9 & 0.0950 & 0.0010 & 0.0941 & \(1.8{\times}10^{-5}\) & 0.0034 & 2.8641 \\
    \bottomrule
  \end{tabular*}
\end{table}

\begin{table}[H]
  \caption{Class-level alpha similarity for pathway level 2 MoA classes.}
  \label{tab:si_alpha_class}
  \centering
  \footnotesize
  \setlength{\tabcolsep}{2pt}
  \renewcommand{\arraystretch}{1.08}
  \begin{tabular*}{\linewidth}{@{\extracolsep{\fill}}p{0.23\linewidth}rrrrp{0.29\linewidth}@{}}
    \toprule
    MoA class & Cond. & Within & Between & Difference & Example drugs \\
    \midrule
    DNA methylation & 4 & 0.7911 & -0.1533 & 0.9445 & Azacitidine; Decitabine; RG108 \\
    Nucleotide analog & 5 & 0.3403 & -0.1200 & 0.4603 & Capecitabine; Carmofur; Fluorouracil \\
    Aurora kinase activity & 9 & 0.2492 & 0.0584 & 0.1908 & Alisertib; Danusertib; Tozasertib \\
    Histone deacetylation & 25 & 0.1021 & 0.0128 & 0.0893 & Abexinostat; Dacinostat; Givinostat \\
    JAK kinase activity & 16 & 0.1125 & 0.0312 & 0.0813 & Baricitinib; Filgotinib; Gandotinib \\
    RTK activity & 15 & 0.0340 & 0.0313 & 0.0028 & Crizotinib; Lapatinib; Cediranib \\
    ADP-ribosylation & 7 & -0.0746 & -0.0099 & -0.0648 & Rucaparib; Veliparib; PJ34 \\
    Nuclear receptor activity & 6 & -0.1244 & -0.0275 & -0.0969 & Fulvestrant; Prednisone; Toremifene \\
    Histone methylation & 6 & -0.1138 & -0.0118 & -0.1020 & Tazemetostat; UNC0631; UNC1999 \\
    \bottomrule
  \end{tabular*}
\end{table}

\begin{table}[H]
  \caption{Nearest-neighbor enrichment of same-class perturbations in \(\alpha(c)\)-space.}
  \label{tab:si_knn}
  \centering
  \footnotesize
  \setlength{\tabcolsep}{3pt}
  \renewcommand{\arraystretch}{1.10}
  \begin{tabular*}{\linewidth}{@{\extracolsep{\fill}}rrrrr@{}}
    \toprule
    \(k\) & Observed same-class & Null mean & Fold enrichment & Permutation \(p\) \\
    \midrule
    1 & 0.4194 & 0.1464 & 2.8641 & 0.0002 \\
    3 & 0.2867 & 0.1464 & 1.9588 & 0.0002 \\
    5 & 0.2323 & 0.1466 & 1.5843 & 0.0002 \\
    8 & 0.1976 & 0.1469 & 1.3455 & 0.0036 \\
    10 & 0.1946 & 0.1469 & 1.3248 & 0.0020 \\
    \bottomrule
  \end{tabular*}
\end{table}

The negative differences in a few MoA classes are also informative. They show that the dictionary does not force every external label to become a compact cluster. Some annotations are broad, some mechanisms are pleiotropic, and some response modes are shared across nominal classes. This is the desired level of evidence for MRFlow: the learned operators are pathway-consistent at aggregate and nearest-neighbor levels, while the method remains flexible enough to share response structure across related pharmacological categories. The result supports mechanism-structured regularization rather than a one-to-one mapping between bases and biological pathways.

Table~\ref{tab:si_alpha_signature} gives a compact view of the same phenomenon from the basis side. Each pathway group is summarized by the two most positive and two most negative mean dictionary activations among the 16 shared affine bases. The signs should not be read as activating or repressing genes directly, since each basis is a latent operator rather than a pathway label. The useful signal is instead relational: different pathway groups recruit different combinations of shared operators, and broad groups such as epigenetic regulation have much smaller average activations than more specialized groups such as nuclear receptor signaling or protein folding and degradation. This pattern is consistent with the design goal of MRFlow, where \(\alpha(c)\) recombines reusable response operators instead of assigning each perturbation to an isolated class-specific model.

\begin{table}[H]
  \caption{Representative pathway-level dictionary activation signatures in SciPlex3. For each pathway, the table reports the two largest and two smallest mean activations among the 16 shared affine bases.}
  \label{tab:si_alpha_signature}
  \centering
  \footnotesize
  \setlength{\tabcolsep}{2pt}
  \renewcommand{\arraystretch}{1.08}
  \begin{tabular*}{\linewidth}{@{\extracolsep{\fill}}p{0.28\linewidth}rp{0.25\linewidth}p{0.25\linewidth}@{}}
    \toprule
    Pathway & Cond. & Largest mean activations & Smallest mean activations \\
    \midrule
    Protein folding \& degradation & 3 & B8=0.86, B11=0.78 & B9=-1.03, B1=-0.74 \\
    Apoptotic regulation & 3 & B13=1.03, B2=0.76 & B12=-0.93, B7=-0.80 \\
    TGF/BMP signaling & 2 & B11=0.68, B13=0.28 & B4=-1.08, B6=-0.85 \\
    Cell cycle regulation & 11 & B4=0.56, B2=0.46 & B1=-0.66, B14=-0.64 \\
    Focal adhesion signaling & 2 & B8=0.57, B4=0.44 & B6=-0.70, B9=-0.49 \\
    Tyrosine kinase signaling & 21 & B8=0.34, B3=0.18 & B13=-0.41, B10=-0.28 \\
    Epigenetic regulation & 39 & B16=0.21, B13=0.19 & B3=-0.31, B12=-0.10 \\
    HIF signaling & 3 & B16=0.32, B9=0.20 & B10=-0.24, B12=-0.23 \\
    JAK/STAT signaling & 20 & B8=0.32, B3=0.21 & B1=-0.27, B10=-0.25 \\
    Metabolic regulation & 4 & B11=0.56, B3=0.51 & B6=-0.61, B15=-0.47 \\
    PKC signaling & 3 & B13=0.40, B2=0.25 & B6=-0.60, B4=-0.28 \\
    Nuclear receptor signaling & 6 & B10=1.43, B1=1.16 & B8=-1.38, B15=-1.12 \\
    DNA damage \& repair & 15 & B10=0.52, B14=0.50 & B16=-0.53, B15=-0.35 \\
    Neuronal signaling & 4 & B9=0.85, B7=0.78 & B11=-0.77, B8=-0.62 \\
    \bottomrule
  \end{tabular*}
\end{table}

The macro-residual decomposition gives a second view of the same architecture. The mechanism-structured affine field is designed to carry the dominant reusable response, whereas the residual field models nonlinear and cell-state-dependent variation. Table~\ref{tab:si_macro_pathway} shows that the macro share is consistently near but above one half across pathway groups, with residual norms remaining substantial. This is exactly the regime intended by the model: the affine dictionary is not bypassed by an unconstrained residual network, and the residual field is not removed by an overly rigid linear transport. The decomposition therefore connects the empirical ablation in Table~\ref{tab:ablation} to an operator-level diagnostic.

\begin{table}[H]
  \caption{Macro and residual velocity decomposition by pathway in SciPlex3.}
  \label{tab:si_macro_pathway}
  \centering
  \footnotesize
  \setlength{\tabcolsep}{2pt}
  \renewcommand{\arraystretch}{1.08}
  \begin{tabular*}{\linewidth}{@{\extracolsep{\fill}}lrrrrr@{}}
    \toprule
    Pathway & Cond. & Macro share & Share SD & Macro norm & Residual norm \\
    \midrule
    TGF/BMP signaling & 2 & 0.5265 & 0.0095 & 3.9287 & 3.5328 \\
    Focal adhesion signaling & 2 & 0.5255 & 0.0098 & 3.9352 & 3.5517 \\
    DNA damage and DNA repair & 15 & 0.5210 & 0.0068 & 3.7837 & 3.4785 \\
    PKC signaling & 3 & 0.5207 & 0.0054 & 3.8088 & 3.5052 \\
    Metabolic regulation & 4 & 0.5205 & 0.0082 & 3.8048 & 3.5050 \\
    JAK/STAT signaling & 20 & 0.5200 & 0.0064 & 3.8543 & 3.5568 \\
    Tyrosine kinase signaling & 21 & 0.5200 & 0.0075 & 3.8336 & 3.5380 \\
    HIF signaling & 3 & 0.5189 & 0.0088 & 3.8290 & 3.5497 \\
    Apoptotic regulation & 3 & 0.5188 & 0.0100 & 3.8185 & 3.5410 \\
    Epigenetic regulation & 39 & 0.5175 & 0.0054 & 3.7700 & 3.5145 \\
    Cell cycle regulation & 11 & 0.5175 & 0.0043 & 3.8396 & 3.5808 \\
    Protein folding and degradation & 3 & 0.5168 & 0.0077 & 3.8623 & 3.6111 \\
    Neuronal signaling & 4 & 0.5163 & 0.0057 & 3.7377 & 3.5009 \\
    Other & 4 & 0.5161 & 0.0076 & 3.7159 & 3.4839 \\
    Nuclear receptor signaling & 6 & 0.5153 & 0.0079 & 3.6422 & 3.4256 \\
    \bottomrule
  \end{tabular*}
\end{table}

Dose-response diagnostics provide a different stress test because dose variation changes response magnitude without changing the drug identity. Table~\ref{tab:si_dose} reports selected four-dose series. These rows should be read as diagnostic provenance for response consistency, not as a global dose-response benchmark. The useful signal is that several series maintain high rank agreement between predicted and observed response strength across the selected doses. At the same time, amplitude ranges differ between measured and predicted responses, which keeps the claim bounded and reinforces why MRFlow is evaluated by multiple raw metrics rather than a single monotonicity score.

\begin{table}[H]
  \caption{Selected dose-response consistency diagnostics. Each drug contributes four dose rows in the source table; the summary reports the shared rank statistics for the selected series.}
  \label{tab:si_dose}
  \centering
  \footnotesize
  \setlength{\tabcolsep}{2pt}
  \renewcommand{\arraystretch}{1.10}
  \begin{tabular*}{\linewidth}{@{\extracolsep{\fill}}llrrrrr@{}}
    \toprule
    Drug & Pathway & Doses & \(\rho_{p,t}\) & \(\rho_{t,d}\) & \(\rho_{p,d}\) & True range \\
    \midrule
    Alisertib & Cell cycle regulation & 4 & 0.8 & 0.4 & -0.2 & 0.1104 \\
    Azacitidine & Epigenetic regulation & 4 & 0.8 & 0.8 & 1.0 & 0.0875 \\
    Dacinostat & Epigenetic regulation & 4 & 0.8 & 1.0 & 0.8 & 0.7595 \\
    G007-LK & TGF/BMP signaling & 4 & 0.8 & 1.0 & 0.8 & 0.0529 \\
    JQ1 & Epigenetic regulation & 4 & 1.0 & 1.0 & 1.0 & 0.2398 \\
    Navitoclax & Apoptotic regulation & 4 & 0.8 & 0.2 & 0.4 & 0.2340 \\
    \bottomrule
  \end{tabular*}
\end{table}

The selected dose series are accompanied by the broader candidate pool in Table~\ref{tab:si_dose_candidates}. This table is important for interpretation because dose-response figures can otherwise look stronger than the underlying evidence. Several candidates, including JQ1, Iniparib, Dacinostat, Navitoclax, Alisertib, Azacitidine, Roxadustat, G007-LK, TMP195, and BMS-911543, have positive rank agreement between predicted and observed response strength. Other candidates have weak or negative rank agreement, especially when the measured dynamic range is small. The mixed pattern is useful rather than damaging: it shows that the appendix is reporting a diagnostic landscape, not cherry-picking a monotone-dose claim. MRFlow is strongest when the perturbation induces a measurable transcriptional response, and the raw-metric evaluation remains the primary evidence for benchmark-level performance.

\begin{table}[H]
  \caption{Candidate dose-response diagnostics across four-dose SciPlex3 drug series. \(\rho_{\mathrm{pred,true}}\) measures rank agreement between predicted and observed response strength across doses.}
  \label{tab:si_dose_candidates}
  \centering
  \footnotesize
  \setlength{\tabcolsep}{2pt}
  \renewcommand{\arraystretch}{1.08}
  \begin{tabular*}{\linewidth}{@{\extracolsep{\fill}}llrrrrr@{}}
    \toprule
    Drug & Pathway & Doses & \(\rho_{p,t}\) & \(\rho_{t,d}\) & \(\rho_{p,d}\) & True range \\
    \midrule
    JQ1 & Epigenetic regulation & 4 & 1.0 & 1.0 & 1.0 & 0.2398 \\
    Iniparib & DNA damage \& repair & 4 & 1.0 & -0.4 & -0.4 & 0.0243 \\
    Dacinostat & Epigenetic regulation & 4 & 0.8 & 1.0 & 0.8 & 0.7595 \\
    Navitoclax & Apoptotic regulation & 4 & 0.8 & 0.2 & 0.4 & 0.2340 \\
    Alisertib & Cell cycle regulation & 4 & 0.8 & 0.4 & -0.2 & 0.1104 \\
    Azacitidine & Epigenetic regulation & 4 & 0.8 & 0.8 & 1.0 & 0.0875 \\
    Roxadustat & HIF signaling & 4 & 0.8 & 0.2 & 0.4 & 0.0537 \\
    G007-LK & TGF/BMP signaling & 4 & 0.8 & 1.0 & 0.8 & 0.0529 \\
    TMP195 & Epigenetic regulation & 4 & 0.8 & 0.8 & 0.4 & 0.0277 \\
    BMS-911543 & JAK/STAT signaling & 4 & 0.8 & -0.8 & -0.6 & 0.0101 \\
    Givinostat & Epigenetic regulation & 4 & 0.4 & 1.0 & 0.4 & 0.7992 \\
    Abexinostat & Epigenetic regulation & 4 & 0.4 & 1.0 & 0.4 & 0.6164 \\
    Nintedanib & Tyrosine kinase signaling & 4 & 0.4 & 0.2 & -0.8 & 0.1065 \\
    AG-14361 & DNA damage \& repair & 4 & 0.4 & 0.8 & 0.0 & 0.0812 \\
    Decitabine & Epigenetic regulation & 4 & 0.4 & 1.0 & 0.4 & 0.0470 \\
    Fasudil HCl & Focal adhesion signaling & 4 & 0.2 & 1.0 & 0.2 & 0.0161 \\
    Rucaparib & DNA damage \& repair & 4 & 0.0 & 1.0 & 0.0 & 0.0292 \\
    Aurora A inh. & Cell cycle regulation & 4 & -0.2 & 1.0 & -0.2 & 0.1015 \\
    Ruxolitinib & JAK/STAT signaling & 4 & -0.4 & 0.4 & -0.4 & 0.0668 \\
    PFI-1 & Epigenetic regulation & 4 & -0.6 & 1.0 & -0.6 & 0.1947 \\
    GSK1070916 & Cell cycle regulation & 4 & -0.8 & 0.8 & -0.6 & 0.0860 \\
    FLLL32 & JAK/STAT signaling & 4 & -1.0 & 1.0 & -1.0 & 0.0170 \\
    \bottomrule
  \end{tabular*}
\end{table}

The qualitative response-gene panels are selected examples, so their selection process must be explicit. Table~\ref{tab:si_dataset_summary} makes this provenance visible across the supplementary panels. Across all eight benchmarks, the selected examples have a same-direction fraction of 1.0 under the selection criterion, but the candidate pool size varies widely, from three available Aissa PC9 conditions to 140 candidate conditions in the Norman and Replogle screens. Table~\ref{tab:si_main_condition_scores} gives additional condition-level provenance for the four HDAC inhibitor panels in Figure~\ref{fig:hdac}. This makes the intended interpretation clear: the panels demonstrate that MRFlow can produce locally inspectable response signatures, while the complete benchmark claims remain those in Tables~\ref{tab:drug_raw} and~\ref{tab:gene_raw}.

\begin{table}[H]
  \caption{Dataset-level provenance for selected supplementary response-gene examples.}
  \label{tab:si_dataset_summary}
  \centering
  \footnotesize
  \setlength{\tabcolsep}{2pt}
  \renewcommand{\arraystretch}{1.08}
  \begin{tabular*}{\linewidth}{@{\extracolsep{\fill}}llrrrrr@{}}
    \toprule
    Dataset & Type & Selected & Same-dir. & Visible signal & Abs. error & Candidate cond. \\
    \midrule
    Adamson 2016 & gene & 4 & 1.0000 & 0.3837 & 0.1127 & 86 \\
    Aissa 2021 PC9 & drug & 3 & 1.0000 & 0.3831 & 0.2073 & 3 \\
    Dixit 2016 & gene & 4 & 1.0000 & 0.2434 & 0.1261 & 19 \\
    Lotfollahi-Theis 2023 & drug & 4 & 1.0000 & 0.4931 & 0.1834 & 30 \\
    McFarland 2020 & drug & 4 & 1.0000 & 0.3718 & 0.1134 & 113 \\
    Norman 2019 & gene & 4 & 1.0000 & 0.1534 & 0.0456 & 140 \\
    Replogle 2022 & gene & 4 & 1.0000 & 0.6401 & 0.1727 & 140 \\
    SciPlex3 & drug & 4 & 1.0000 & 0.3152 & 0.1956 & 114 \\
    \bottomrule
  \end{tabular*}
\end{table}

\begin{table}[H]
  \caption{Condition-level provenance for the HDAC inhibitor response-gene panels in Figure~\ref{fig:hdac}.}
  \label{tab:si_main_condition_scores}
  \centering
  \footnotesize
  \setlength{\tabcolsep}{2pt}
  \renewcommand{\arraystretch}{1.08}
  \begin{tabular*}{\linewidth}{@{\extracolsep{\fill}}lllrrr@{}}
    \toprule
    Label & Drug name & Condition & Cells & DE20 Pearson & Displayed direction match \\
    \midrule
    Dacinostat & Dacinostat (LAQ824) & 1000 nM, A549 & 252 & 0.8959 & 15/15 \\
    Abexinostat & Abexinostat (PCI-24781) & 1000 nM, A549 & 413 & 0.8497 & 15/15 \\
    Pracinostat & Pracinostat (SB939) & 1000 nM, A549 & 456 & 0.6576 & 15/15 \\
    Givinostat & Givinostat (ITF2357) & 1000 nM, A549 & 219 & 0.4800 & 15/15 \\
    \bottomrule
  \end{tabular*}
\end{table}

Figure~\ref{fig:structured} in the main text is a compact diagnostic figure, so Tables~\ref{tab:si_case_components} and~\ref{tab:si_case_genes} expose the tabular evidence behind the case-level panels. The component table asks whether a representative response is dominated by one basis or distributed across the structured affine dictionary. The largest single component is the bias term, followed by multiple shared bases, while the low-rank correction contributes a small fraction. This is consistent with the intended decomposition: the dictionary supplies reusable operator structure, the bias captures a shared affine offset, and the low-rank term gives condition-specific flexibility without replacing the shared basis.

\begin{table}[H]
  \caption{Representative Figure~\ref{fig:structured} case decomposition by affine operator component. Fractions are computed within the displayed case.}
  \label{tab:si_case_components}
  \centering
  \footnotesize
  \setlength{\tabcolsep}{3pt}
  \renewcommand{\arraystretch}{1.10}
  \begin{tabular*}{\linewidth}{@{\extracolsep{\fill}}lrrrlrr@{}}
    \toprule
    Component & Value & Fraction & \(\alpha\) & Component & Value & Fraction \\
    \midrule
    Bias & 5.3610 & 0.2760 &  & B11 & 1.5351 & 0.0790 \\
    B1 & 2.0384 & 0.1049 & -1.0145 & B12 & 1.2531 & 0.0645 \\
    B4 & 1.2143 & 0.0625 & -0.4890 & B10 & 1.2013 & 0.0618 \\
    B14 & 1.1755 & 0.0605 & -1.0878 & B5 & 1.1158 & 0.0574 \\
    B7 & 1.0273 & 0.0529 & -0.9672 & B6 & 0.8725 & 0.0449 \\
    B9 & 0.6588 & 0.0339 & -0.4830 & B13 & 0.6448 & 0.0332 \\
    B8 & 0.3839 & 0.0198 & 0.4792 & Low-rank & 0.2797 & 0.0144 \\
    B15 & 0.2771 & 0.0143 & 0.1111 & B16 & 0.2444 & 0.0126 \\
    B2 & 0.0721 & 0.0037 & 0.0725 & B3 & 0.0694 & 0.0036 \\
    \bottomrule
  \end{tabular*}
\end{table}

The case response-gene table gives a gene-level sanity check on the same operator decomposition. All displayed genes preserve the down-regulated direction in this example, while absolute errors vary by gene. The table therefore supports a restrained interpretation of the case panel: MRFlow can produce an internally consistent chain from condition embedding to operator mixture to signed gene shifts for selected high-signal cases, but this example is not used as a substitute for the aggregate benchmark tables. Its role is to show that the mechanism-structured latent transport has inspectable local outputs.

\begin{table}[H]
  \caption{Representative Figure~\ref{fig:structured} case response-gene signature. Shifts are signed relative to the source population.}
  \label{tab:si_case_genes}
  \centering
  \footnotesize
  \setlength{\tabcolsep}{3pt}
  \renewcommand{\arraystretch}{1.10}
  \begin{tabular*}{\linewidth}{@{\extracolsep{\fill}}lrrlr@{}}
    \toprule
    Gene & True shift & Pred. shift & Direction & Abs. error \\
    \midrule
    AKR1B10 & -1.7022 & -1.9314 & down & 0.2292 \\
    HNF1A-AS1 & -0.9813 & -0.9879 & down & 0.0066 \\
    CACNA1D & -0.6709 & -0.7485 & down & 0.0776 \\
    CPLX2 & -0.5835 & -0.6793 & down & 0.0957 \\
    CP & -0.4373 & -0.4643 & down & 0.0270 \\
    TM4SF20 & -0.3952 & -0.3874 & down & 0.0078 \\
    CLDN2 & -0.3046 & -0.3478 & down & 0.0431 \\
    FCGBP & -0.2920 & -0.3332 & down & 0.0412 \\
    RARRES1 & -0.2197 & -0.2357 & down & 0.0160 \\
    TMEM156 & -0.2120 & -0.2451 & down & 0.0331 \\
    EDARADD & -0.1573 & -0.1945 & down & 0.0373 \\
    SLC25A21 & -0.1105 & -0.1160 & down & 0.0055 \\
    \bottomrule
  \end{tabular*}
\end{table}

Several larger Table-SI files are better treated as machine-readable provenance than as PDF tables. The condition-level macro-residual decomposition contains 141 canonical SciPlex3 conditions, the canonical alpha activation table contains 141 rows with 16 dictionary-basis coordinates, the selected supplementary response-gene table contains 620 gene-level rows, and the full candidate-condition score table contains 645 rows. We therefore cite them as supplementary data rather than typesetting them in full. Their role is to make the qualitative and mechanism-structured diagnostics reproducible: a reader can trace from an appendix figure to the selected condition, from the selected condition to its candidate pool, and from a pathway-level diagnostic to the condition-level activation values used to compute it.

\section{Supplementary Matched Response-Gene Shift Examples}
\label{app:matched_shift_examples}

To complement the aggregate metrics in the main text, we provide matched response-gene shift visualizations across all eight benchmark datasets. Each panel corresponds to one held-out perturbation condition and displays 20 response genes, with 10 upregulated genes and 10 downregulated genes. For each gene, the observed signed expression shift is compared with the MRFlow-predicted signed shift, where both shifts are computed relative to the corresponding control or source population.

These panels are generated from processed AnnData files and MRFlow checkpoint inference; no simulated values are used. The examples are selected as high-matching qualitative cases using a balanced criterion that favors direction agreement, visible response magnitude, and low absolute error. They should therefore be interpreted as qualitative model-inspection panels rather than unbiased estimates of global performance. The aggregate performance claims remain based on the raw metrics reported in Tables~\ref{tab:drug_raw} and~\ref{tab:gene_raw}.

These examples provide interpretability at the level that a biologist or drug-response analyst would inspect after reading the aggregate tables. A distributional model may score well on global metrics but still produce response-gene shifts that are difficult to interpret condition by condition. These panels ask a local question: when a held-out condition has a clear signed response signature, does the predicted signature preserve the direction and relative size of the response genes? The examples do not replace quantitative evaluation, but they make the model output more auditable.

The selected genes are displayed as signed shifts relative to control rather than as raw expression values. This presentation aligns the panels with the DE20 metrics in the main text and with the biological question of perturbation response. It also avoids conflating baseline expression level with induced change. For AIVC use, the induced change is often the actionable object: it indicates which genes are predicted to rise or fall under an intervention and whether the predicted response agrees with the observed perturbation signature.

\begin{figure}[H]
  \centering
  \includegraphics[width=0.82\linewidth]{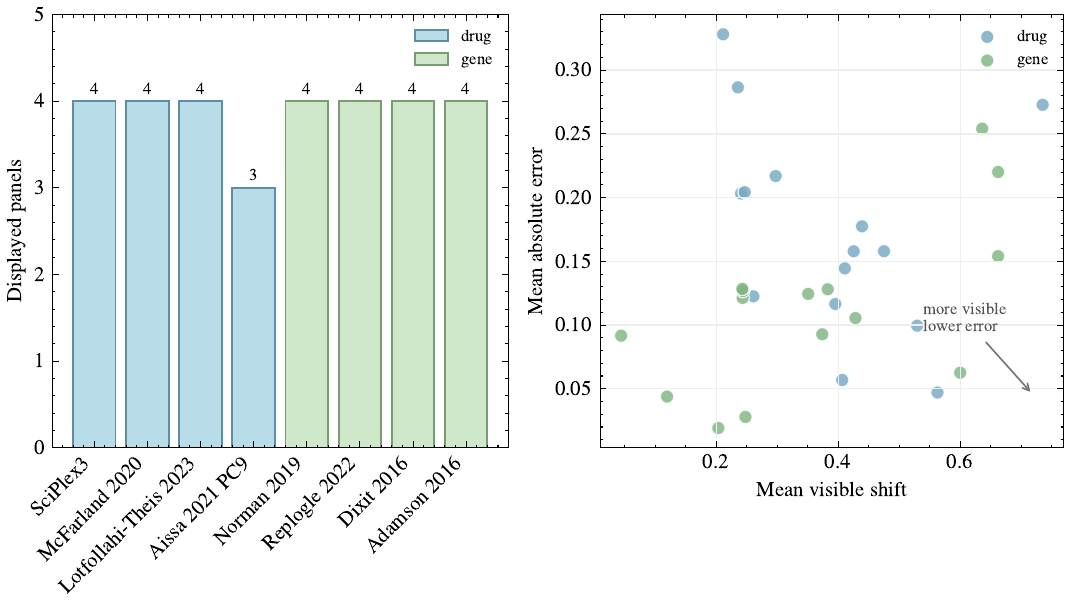}
  \caption{Coverage of selected response-gene shift examples. The supplementary panels cover four drug perturbation benchmarks and four genetic perturbation benchmarks, with three real conditions for Aissa 2021 PC9 because the processed benchmark contains only three non-control drug conditions.}
  \label{fig:app_qual_coverage}
\end{figure}

\begin{figure}[H]
  \centering
  \includegraphics[width=0.82\linewidth]{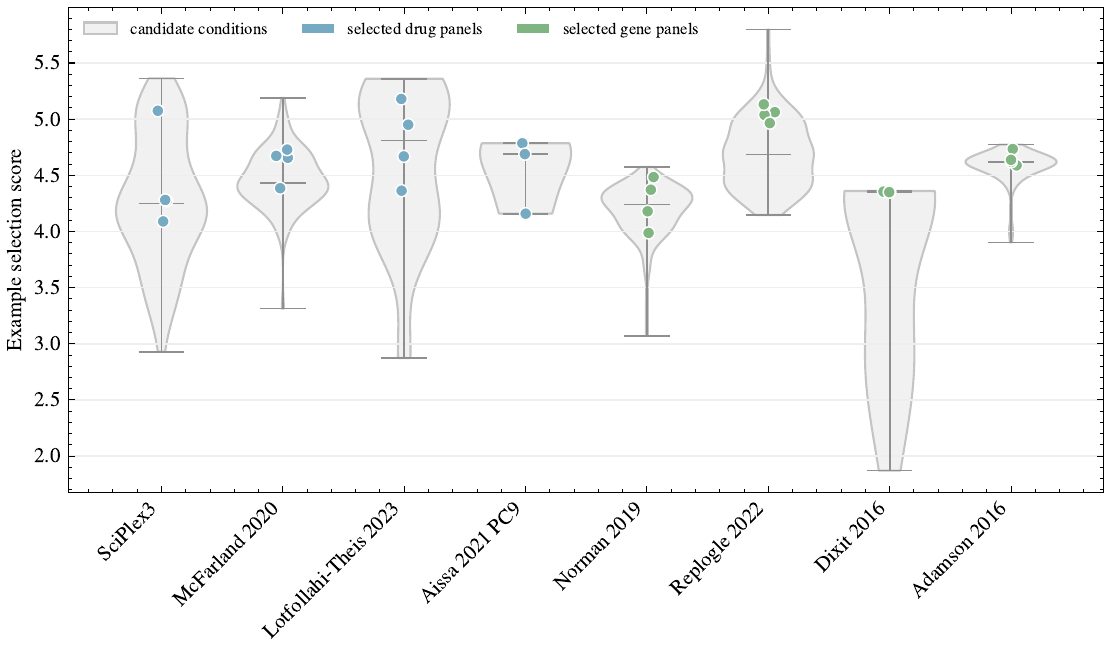}
  \caption{Selection protocol for matched response-gene shift panels. Candidate held-out perturbation conditions are scored by direction agreement, visible response magnitude, and absolute error, then selected as qualitative high-matching examples for inspection.}
  \label{fig:app_selection_protocol}
\end{figure}

Figures~\ref{fig:app_qual_coverage} and~\ref{fig:app_selection_protocol} document the coverage and selection protocol for these supplementary panels. The Aissa 2021 PC9 processed benchmark contains only three non-control perturbation conditions, so three real examples are shown for that dataset rather than adding an artificial fourth panel.

The qualitative panels serve a different purpose from the tables. The raw metrics establish dataset-level performance, while these panels ask whether the model output is readable at the gene-signature level for concrete held-out conditions. The coverage and selection figures guard against overinterpretation by labeling examples as selected high-matching cases. The complete raw-metric evaluation in the main tables remains the primary evidence for average performance.

\newpage
\input{checklist.tex}

\end{document}

%% file: checklist.tex
\section*{NeurIPS Paper Checklist}

\begin{enumerate}

\item {\bf Claims}
    \item[] Question: Do the main claims made in the abstract and introduction accurately reflect the paper's contributions and scope?
    \item[] Answer: \answerYes{}.
    \item[] Justification: The claims are stated in the abstract and Introduction, and they are limited to conditional distribution transport, mechanism-residual modeling, and empirical performance on the reported perturbation benchmarks.

\item {\bf Limitations}
    \item[] Question: Does the paper discuss the limitations of the work performed by the authors?
    \item[] Answer: \answerYes{}.
\item[] Justification: The paper explicitly bounds its claims to held-out condition prediction on the reported perturbation benchmarks, clarifies that MRFlow is evaluated per dataset rather than as a jointly trained cross-dataset model, and notes that broader dose programs, perturbation modalities, and experimental validation remain directions for future work.

\item {\bf Theory assumptions and proofs}
    \item[] Question: For each theoretical result, does the paper provide the full set of assumptions and a complete and correct proof?
    \item[] Answer: \answerNA{}.
    \item[] Justification: The paper introduces MRFlow as a model contribution, with innovations in perturbation-conditioned latent distribution transport, a mechanism-structured affine dictionary, residual velocity modeling, and OT-guided pairing for unpaired single-cell populations. These contributions are defined mathematically but are not presented as formal theorems requiring proof.

\item {\bf Experimental result reproducibility}
    \item[] Question: Does the paper fully disclose all the information needed to reproduce the main experimental results of the paper to the extent that it affects the main claims and conclusions of the paper, regardless of whether the code and data are provided or not?
    \item[] Answer: \answerYes{}.
\item[] Justification: The paper specifies the datasets, baselines, raw metrics, composite-score definition, low-cost pairing implementation, model components, training objective, and evaluation aggregation used for the main claims.

\item {\bf Open access to data and code}
    \item[] Question: Does the paper provide open access to the data and code, with sufficient instructions to faithfully reproduce the main experimental results, as described in supplemental material?
    \item[] Answer: \answerNo{}.
    \item[] Justification: The current submission uses publicly cited datasets and fully describes the method and evaluation protocol, but the anonymized code and reproduction package are not yet attached. Upon acceptance, the authors will publicly release the full MRFlow code, preprocessing scripts, configuration files, and data access or processed-data instructions needed to reproduce the reported experiments.

\item {\bf Experimental setting/details}
    \item[] Question: Does the paper specify all the training and test details necessary to understand the results?
    \item[] Answer: \answerYes{}.
\item[] Justification: The Experiments section and appendix identify datasets, held-out condition split assumptions, baselines, metrics, composite scoring, OT-guided pairing, training objective terms, preprocessing notes, software versions, hardware, and training-time range.

\item {\bf Experiment statistical significance}
    \item[] Question: Does the paper report error bars suitably and correctly defined or other appropriate information about the statistical significance of the experiments?
    \item[] Answer: \answerYes{}.
    \item[] Justification: The paper reports complete raw metrics for each dataset-method pair and additional raw-metric rank diagnostics across dataset-metric cells. These diagnostics provide fixed-split benchmark evidence for the comparative claims.

\item {\bf Experiments compute resources}
    \item[] Question: For each experiment, does the paper provide sufficient information on the computer resources needed to reproduce the experiments?
    \item[] Answer: \answerYes{}.
    \item[] Justification: Appendix~\ref{app:compute_impact} reports the GPU, memory, CPU, operating system, Python version, PyTorch version, CUDA version, and per-dataset training-time range used for the experiments.

\item {\bf Code of ethics}
    \item[] Question: Does the research conducted in the paper conform, in every respect, with the NeurIPS Code of Ethics?
    \item[] Answer: \answerYes{}.
    \item[] Justification: The work uses previously published single-cell perturbation datasets and develops a predictive model for biological research without involving new human-subject data collection.

\item {\bf Broader impacts}
    \item[] Question: Does the paper discuss both potential positive societal impacts and negative societal impacts of the work performed?
    \item[] Answer: \answerYes{}.
    \item[] Justification: Appendix~\ref{app:compute_impact} discusses potential benefits for perturbation prioritization and drug-response analysis, as well as risks from over-interpreting in-silico predictions without experimental validation.

\item {\bf Safeguards}
    \item[] Question: Does the paper describe safeguards that have been put in place for responsible release of data or models that have a high risk for misuse?
    \item[] Answer: \answerNA{}.
    \item[] Justification: The paper does not release a high-risk generative model for open-ended content or scraped personal data.

\item {\bf Licenses for existing assets}
    \item[] Question: Are the creators or original owners of assets used in the paper properly credited and are the license and terms of use explicitly mentioned and properly respected?
    \item[] Answer: \answerNo{}.
    \item[] Justification: The datasets and prior models are cited, and Appendix~\ref{app:baseline} states that all datasets and baseline implementations come from public papers or official open-source releases. The final reproduction package will follow the original licenses and access terms, but the current draft does not yet list every dataset and software license individually.

\item {\bf New assets}
    \item[] Question: Are new assets introduced in the paper well documented and is the documentation provided alongside the assets?
    \item[] Answer: \answerNA{}.
    \item[] Justification: The paper does not introduce a new biological dataset, benchmark asset, or public model checkpoint as part of the current submission. The planned code release is described under the open-access item above.

\item {\bf Crowdsourcing and research with human subjects}
    \item[] Question: For crowdsourcing experiments and research with human subjects, does the paper include the full text of instructions given to participants and screenshots, if applicable, as well as details about compensation?
    \item[] Answer: \answerNA{}.
    \item[] Justification: The work does not involve crowdsourcing or new experiments with human subjects.

\item {\bf Institutional review board approvals or equivalent for research with human subjects}
    \item[] Question: Does the paper describe potential risks incurred by study participants, whether such risks were disclosed to the subjects, and whether Institutional Review Board approvals or an equivalent approval were obtained?
    \item[] Answer: \answerNA{}.
    \item[] Justification: The work analyzes previously published cell-line perturbation datasets and does not recruit human participants.

\item {\bf Declaration of LLM usage}
    \item[] Question: Does the paper describe the usage of LLMs if it is an important, original, or non-standard component of the core methods in this research?
    \item[] Answer: \answerNA{}.
    \item[] Justification: LLMs are not part of the MRFlow model architecture, training objective, or experimental evaluation described in the paper.

\end{enumerate}